%% file: thesis.tex
\documentclass[10pt,twocolumn,letterpaper]{article}

\usepackage[letterpaper,margin=1in]{geometry}
\usepackage[T1]{fontenc}
\usepackage[utf8]{inputenc}
\usepackage{times}
\usepackage{amsmath}
\usepackage{amssymb}
\usepackage{booktabs}
\usepackage{graphicx}
\graphicspath{{figures/}}
\usepackage{array}
\usepackage{url}
\usepackage{microtype}
\usepackage[numbers,sort&compress]{natbib}
\usepackage{pgfplots}
\pgfplotsset{compat=1.16}
\usepgfplotslibrary{fillbetween}
\usepackage[hidelinks]{hyperref}

\input{generated/results_macros.tex}
\input{generated/phase1_macros.tex}
\input{generated/phase3_macros.tex}

\pgfplotsset{
  pzstyle/.style={
    width=0.80\columnwidth, height=4.4cm, scale only axis,
    tick label style={font=\scriptsize},
    label style={font=\scriptsize},
    legend style={font=\scriptsize, draw=none, fill=none,
                  cells={anchor=west}, inner sep=1pt},
    grid=major, grid style={gray!20, very thin},
    every axis plot/.append style={very thick},
  }
}

\makeatletter
\renewcommand\section{\@startsection{section}{1}{\z@}%
  {-2.0ex \@plus -0.6ex \@minus -.2ex}{1.0ex \@plus .2ex}%
  {\normalfont\large\bfseries}}
\renewcommand\subsection{\@startsection{subsection}{2}{\z@}%
  {-1.6ex \@plus -0.5ex \@minus -.2ex}{0.7ex \@plus .2ex}%
  {\normalfont\normalsize\bfseries}}
\makeatother

\newcommand{\BA}{\mathrm{BA}}
\newcommand{\nuis}{\mathrm{nuis}}
\newcommand{\Real}{\mathbb{R}}

\newcommand{\illus}{\textit{Illustrative.}}

\title{\bf Excess Separability: Nuisance-Controlled Residual-Stream\\ Probing for Benchmark Contamination Detection}

\author{Florian Braun \\
  \texttt{\href{mailto:braunf25@proton.me}{braunf25@proton.me}} \quad
  \texttt{\href{https://braunf.com/}{braunf.com}}}
\date{}

\newcommand\blfootnote[1]{%
  \begingroup
  \renewcommand\thefootnote{}\footnote{#1}%
  \addtocounter{footnote}{-1}%
  \endgroup}

\begin{document}
\maketitle

\blfootnote{Code, item-set builders, the Phase 0c validation study and the
artefact manifest are at
\href{https://github.com/mabushi-lab/residual-stream-contamination-probing}%
{github.com/mabushi-lab/residual-stream-contamination-probing}. The exact
snapshot these results were produced from is archived at
\href{https://doi.org/10.5281/zenodo.21855510}{doi:10.5281/zenodo.21855510}.
Upstream dataset revisions are pinned, so the item sets rebuild to the hashes
recorded in the manifest. The Pile extracts are rebuilt rather than
redistributed; GSM1k is not redistributable at all.}

\begin{abstract}
\noindent
Benchmark contamination is diagnosed today with $n$-gram overlap, with
likelihood-based membership inference, or with canary strings, and each needs
something usually unavailable: the training corpus, a well-chosen test
statistic, or foresight at dataset release. A recent alternative reads
contamination off a linear probe on the model's internal activations. We show
that the natural way to do this does not work, specify one that survives
measurement, and then find that the correction which makes it work carries
more variance than the null it is tested against.

The protocol reports a zero-sum contrast on the depth profile of probe
accuracy, recentred on a level-matched placebo baseline and tested against a
label-permutation null, with the reference set twice the size of the suspect
set. Each of those four choices replaces a simpler alternative that we measured
and rejected. Reporting the level of excess separability rather than its shape
makes the false positive rate track the size of the analyst's own control set,
from 0.03 to 0.99 under a true null. Contrasting against a flat depth profile
fails in both directions, rejecting a true null 0.72 of the time when surface
decodability rises with depth and losing all power when it falls, which is what
a residual stream with accumulating noise produces. An item bootstrap holds the
fitted probe fixed and rejects up to 0.09 of the time where a permutation null
that refits it holds 0.02. A half-size baseline triples the error rate. The corrected protocol holds the nominal rate across
all of these, under two deliberately dissimilar simulators, and reaches 0.8
power at a separation of roughly one to two accuracy points between the
embedding layer and the profile peak.

We then run the protocol on real transformers, and it fails a test the
simulations did not pose. Baseline depth profiles are measurably not flat,
spanning up to \PoneWkSpanHi{} accuracy points on a temporal split, and their
non-flatness tracks the surface difference between the two item sets
(correlation \PoneCorr{} over \PoneN{} audits), so the correction is largest
exactly where it is needed. But the correction is an estimate, and the
permutation null holds it fixed. Re-estimating it across split seeds on
\PthreeN{} audits of deliberately contaminated checkpoints gives a standard
deviation \PthreeRatioLo{} to \PthreeRatioHi{} times the null's own, in every
arm: the quantity subtracted to remove a bias is more variable than the
quantity it is subtracted from. The one nominally significant result in this
paper, $p=\PthreeBlockedP{}$, becomes $p=\PthreeBlockedPinfl{}$ once that
variance is propagated, and the protocol issues no verdict. The simulations
missed this because their surface key is the covariate driving item variation;
on real text it is a proxy, and degrading key quality in simulation reproduces
the effect. We therefore add a required companion measurement and a widened
null, and report that no arm shows contamination: all \PonePileNulls{}
well-matched Pile arms are null, the temporal split is refused for failing
exchangeability, and the contaminated checkpoints are null at both duplication
counts tested. Whether transformers carry a familiarity direction remains
open. We release the implementation, the test suite, the validation study and
the audits.
\end{abstract}

\section{Introduction}

A benchmark score is a claim about capability. Contamination breaks the claim by making the score partly a report of what the model has memorised. The problem is old, was documented at scale in the GPT-3 report \cite{brown2020gpt3}, and has grown worse as training corpora have grown and as their contents have stopped being disclosed \cite{sainz2023trouble,balloccu2024leak}.

The detection tools available today divide into three families, each blocked by a different missing precondition. Substring and $n$-gram overlap search requires the training corpus, which for most released models does not exist publicly \cite{dodge2021documenting,elazar2024wimbd}. Likelihood-based membership inference requires only the model, but fixes a scalar test statistic in advance (loss, minimum-$k$\% token log-probability, a reference-model ratio) and inherits whatever that statistic fails to capture \cite{shi2024minkprob,zhang2025minkpp,mattern2023neighbourhood}; evaluated under splits that are genuinely exchangeable, this family performs close to chance on pretraining-scale data \cite{duan2024mia}. Canary insertion works well but has to be planned before the benchmark is published \cite{carlini2019secretsharer}, which does nothing for the benchmarks already in use.

A fourth option has appeared recently. \citet{liu2024probing} train a linear probe on a model's internal activations and report state-of-the-art AUC on WikiMIA and on a new arXiv-based benchmark. The motivation is sound: if memorisation leaves any trace in the model, it should be visible in the residual stream before it is visible in a hand-picked summary of the output distribution, and a probe learns the relevant direction instead of assuming it.

The difficulty is that a probe trained to separate two item sets will use anything that separates them. \citet{das2024blind} show that on several widely used membership-inference benchmarks, including WikiMIA, a classifier with no access to the target model at all reaches high AUC, because the member and non-member sets were split by publication date and therefore differ in topic and style. Any probe evaluated on such a split reports the sum of two quantities, memorisation and distribution shift, and the split's construction guarantees the second is large. The strongest evidence in \citeauthor{liu2024probing}'s own paper is consistent with this reading: on the deliberately contaminated checkpoints of \citet{oren2024proving}, where the contaminated and clean item sets come from the same dataset and the split is not temporal, their probe reaches 54.0 and 51.9 AUC at duplication counts of 1 and 2, against a 50.0 chance baseline.

This paper takes the position that the probe is the right instrument and the reported quantity is the wrong one. We propose \textbf{Residual-Stream Contamination Probing} (RSCP), a protocol whose output is the probe's advantage \emph{over a declared nuisance family}, tested against a permutation null, and calibrated against known exposure counts.

\paragraph{Contributions.}
\begin{enumerate}\itemsep2pt
\item An exchangeability requirement on the reference set, stated as a
precondition rather than a caveat, with three constructions that satisfy it
and a diagnosis of which existing evaluation sets do not (\S\ref{sec:exchange}).
\item An extraction rule: activations are read at the item prefix, because a
state computed over the answer lets the detector read the model's own
competence, which is the effect contamination is supposed to explain
(\S\ref{sec:extraction}).
\item A statistic that is a contrast on the depth profile rather than a level,
which cancels the analyst's control set algebraically (\S\ref{sec:phase0}, C1).
\item A level-matched placebo baseline, which removes the assumption that
surface decodability is flat in depth. This is the paper's main technical
contribution: the same construction that vetoes an unusable audit also
supplies the baseline that makes a usable one correct (\S\ref{sec:phase0}, C2).
\item A permutation null on a label-free smoother, which refits the probe and
so captures the variance an item bootstrap misses, at one matrix-vector
product per draw (\S\ref{sec:phase0}, C3).
\item A validation study under two structurally dissimilar simulators that
falsified three of our own earlier design choices, together with the reasoning
that replaced them (\S\ref{sec:phase0}).
\item Phase 1 of the empirical programme, executed: \PoneN{} audits on real
Pythia models establishing that baseline depth profiles are not flat, that
their non-flatness scales with the surface difference, and that the protocol's
negative control behaves (\S\ref{sec:phase1}).
\item A runnable implementation, a property test suite, and a build that
regenerates every number in this paper from scratch in about ten minutes on a
laptop (\S\ref{sec:repro}).
\item A pre-registered follow-on programme with falsification criteria and a
compute budget on the order of 100 GPU-hours (\S\ref{sec:programme}).
\end{enumerate}

One boundary is worth drawing at the outset. Section~\ref{sec:phase0} reports measurements of a statistical procedure on synthetic activations; Section~\ref{sec:phase1} reports measurements on real models. Neither is evidence that a transformer carries a familiarity direction, for reasons \S\ref{sec:phase1} sets out. Everywhere else, every number is either a published measurement with a citation or a synthetic value marked \emph{illustrative} to show the shape of the output rather than to anticipate a result.

\section{Background and Related Work}

\subsection{What contamination is, and what makes it hard to see}

Contamination is not a single phenomenon. Verbatim inclusion of a test item in the pretraining corpus, inclusion of a paraphrase, inclusion of the item's source document without the question-answer framing, and inclusion of a downstream discussion of the benchmark all inflate scores by different mechanisms and to different degrees. \citet{magar2022exploitation} separate memorisation, meaning the model has stored the item, from exploitation, meaning the stored item actually changes the answer, and show the two come apart: a model can memorise contaminated items without exploiting them, and the size of the exploitation effect depends on the fine-tuning regime. \citet{yang2023rephrased} show that paraphrased benchmark items evade $n$-gram detection while still inflating scores, which rules out overlap search as a complete solution even when the corpus is available.

The corpus is usually not available, and where it is, overlap search has free parameters that matter: the GPT-3 analysis \cite{brown2020gpt3} and the C4 documentation study \cite{dodge2021documenting} use different $n$-gram lengths and normalisation and are not directly comparable. Corpus-search tooling has improved \cite{elazar2024wimbd,soldaini2024dolma}, but only for models whose corpora are published.

Corpus-free methods therefore carry the field. \citet{sainz2023trouble} prompt models to reproduce dataset splits verbatim. \citet{golchin2024timetravel} use guided instructions and a quiz format. \citet{deng2024investigating} exploit the model's ability to reconstruct masked incorrect options. \citet{li2024task} use release dates as a natural experiment. \citet{chang2023speak} infer which books a model has seen from cloze accuracy on passages, which is the same idea applied to copyrighted text rather than to benchmarks. \citet{oren2024proving} give the most statistically careful of these: benchmark items are exchangeable under permutation of their order, so a model that assigns higher log-likelihood to the canonical ordering than to shuffled orderings has seen the canonical ordering, and this yields a valid $p$-value with no distributional assumptions. Its limitation is that it detects contamination of the dataset \emph{as a file}, so it is blind to contamination that arrived through paraphrase, through a different serialisation, or one item at a time.

\subsection{Membership inference and its exchangeability problem}

Membership inference asks whether a specific example was in the training set \cite{shokri2017mia,yeom2018privacy}. The strongest formulations calibrate against a reference distribution over models \cite{carlini2022firstprinciples}, which is unavailable at pretraining scale, so LLM-oriented work uses reference-free statistics: the loss itself \cite{yeom2018privacy}, likelihood ratios adapted from the masked-language-model setting \cite{mireshghallah2022quantifying}, comparison against synthetically perturbed neighbours \cite{mattern2023neighbourhood}, the mean log-probability of the lowest-probability $k$\% of tokens \cite{shi2024minkprob}, and a normalised variant motivated by the observation that training samples sit at local maxima of the modelled distribution along each input dimension \cite{zhang2025minkpp}.

\citet{duan2024mia} evaluate this family across Pythia models from 160M to 12B on Pile train versus Pile test and find performance barely above chance, attributing it to a very large corpus seen close to once and a member boundary made fuzzy by near-duplicates. \citet{das2024blind} sharpen the consequence: where the split is temporal, a blind classifier that never queries the target model achieves high AUC, so high AUC is not evidence of memorisation detection. Together these say that reported membership-inference performance on LLMs is largely an artefact of how the two sets were built, and that under exchangeable splits the true signal is small.

\citet{maini2024datasetinference} draw the constructive conclusion. Individual membership is close to undetectable, but aggregating a weak per-item signal over a whole dataset is a different and much easier statistical problem. Their dataset inference procedure combines several per-item statistics and tests at the set level. RSCP sits in this tradition: it is a two-sample test on sets, not a classifier deployed on individual items, and we frame it that way throughout.

\subsection{Probing classifiers: what a probe licenses}

Linear probes read a property off intermediate representations \cite{alain2016probes}, and the interpretive difficulty has been mapped carefully. \citet{hewitt2019control} show that probe accuracy confounds what the representation encodes with what the probe can learn, and introduce control tasks with randomised labels together with selectivity. \citet{voita2020mdl} and \citet{pimentel2020info} recast probing information-theoretically and note that a sufficiently expressive probe recovers any property computable from the input, so accuracy alone is uninformative without a baseline. \citet{belinkov2022probing} surveys the consensus: probe accuracy means something only relative to a declared control, and a negative result is weak evidence of absence, showing only that the property is not \emph{linearly} decodable at the chosen extraction point.

RSCP inherits all of this. The controls we require, in \S\ref{sec:nuisance}, are the direct descendants of \citeauthor{hewitt2019control}'s control tasks, extended in one direction that their setting did not need: because the property under test here is defined by a difference between two item \emph{sets}, the relevant control has two parts: a randomised-label task, and the strongest classifier buildable from the surface difference between the sets.

\subsection{Why one might expect a familiarity direction}

The mechanistic prior for RSCP is that models carry linearly decodable signals about their own epistemic state. Models have calibrated internal representations of whether they will answer correctly \cite{kadavath2022know}. Truth of a declarative statement is linearly decodable from residual-stream activations across several datasets, with the recovered directions transferring between datasets \cite{marks2024geometry}, can be recovered without labels from consistency constraints \cite{burns2023dlk}, and separates statements the model itself is asserting falsely \cite{azaria2023internal}. Sparse probing finds context and domain features represented in individual neurons \cite{gurnee2023haystack}. Representation-engineering work shows that such directions are not merely correlational, since adding them to the residual stream changes behaviour in the predicted direction \cite{zou2023repe,rimsky2024caa}.

None of this establishes that a \emph{familiarity} direction exists. It establishes that the hypothesis is of a type that has been confirmed before, which is the right standing for a proposal paper. It also sets up the most informative possible negative result, discussed in \S\ref{sec:discussion}: if excess separability is near zero on models with verified single-exposure contamination, then representational familiarity has an exposure floor, and the field should stop expecting activation-based detectors to catch low-duplication leakage.

\begin{figure*}[t]
\centering
\includegraphics[width=0.98\textwidth]{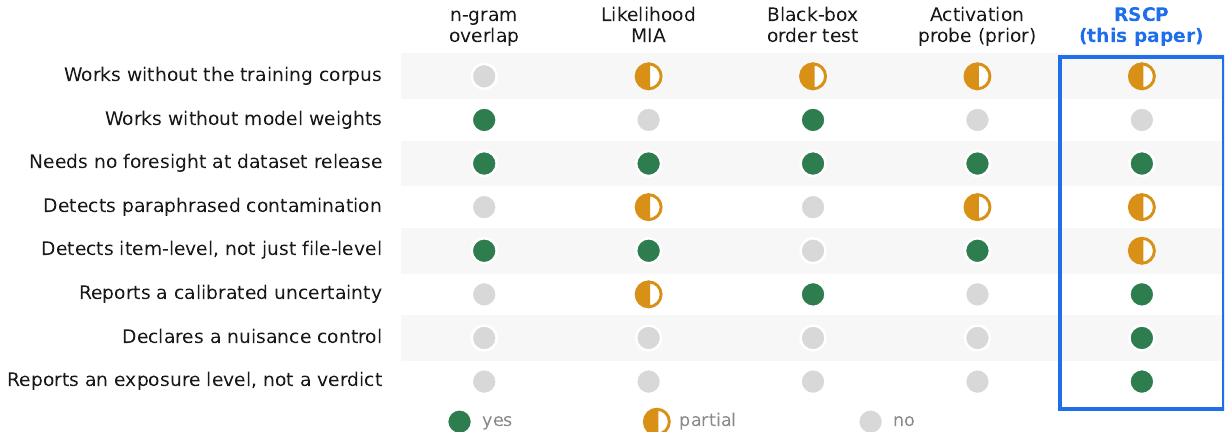}
\caption{Preconditions and capabilities of the four existing families and of RSCP. No column dominates: the black-box order test is the only option for a closed model, and overlap search is the only one that says which documents leaked. The bottom two rows are what this paper adds, and the second row is what it gives up. Judgements, not measurements.}
\label{fig:methods}
\end{figure*}

Figure~\ref{fig:methods} places these families side by side. What it makes visible is that nothing in the existing set declares a nuisance control or reports contamination as a graded quantity rather than a verdict.

\subsection{The immediate predecessor}

\citet{liu2024probing} is the closest work and the one RSCP is designed to correct. Their pipeline fine-tunes the target model on a set of documents to produce a proxy model that has demonstrably memorised them, trains a logistic probe on last-token activations from the proxy at a validation-selected layer, and applies that probe to the original target model. On WikiMIA with Pythia-2.8B they report 69.8 AUC against 65.5 for the best baseline, and on their ArxivMIA benchmark with OpenLLaMA-13B they report 60.3 against 55.9.

Four features of that design are what RSCP changes. Both evaluation sets are split temporally: WikiMIA members are pre-cutoff Wikipedia event text, and ArxivMIA members come from the RedPajama arXiv subset against abstracts published after 2024. These are the splits \citet{das2024blind} show to be separable blind, so the reported AUCs upper-bound rather than estimate the memorisation signal. The extraction point is the last token of a template that wraps the full sample and appends an evaluative question, reported as ``Here is a statement: [SAMPLE] Is the above statement correct? Answer:''. That state is downstream of the entire item and the prompt explicitly asks the model to evaluate it; for benchmark contamination the consequences are severe, because competence on the item is precisely what contamination inflates (\S\ref{sec:extraction}). The layer is selected on a validation set and reported without correction for the search. And the probe transfers from a fine-tuned proxy to the target model, which assumes that the direction separating members after supervised fine-tuning is the direction pretraining memorisation writes to; that assumption is substantive and is not tested.

Their result on the deliberately contaminated checkpoints of \citet{oren2024proving}, where the split is not temporal, is the honest calibration point for the whole approach and we return to it in \S\ref{sec:we2}.

\section{Method}
\label{sec:method}

\begin{figure*}[t]
\centering
\includegraphics[width=0.98\textwidth]{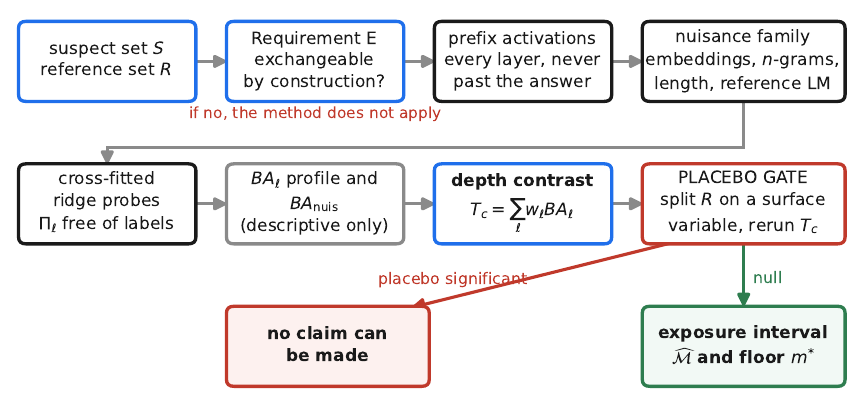}
\caption{The protocol end to end. Two of the boxes can terminate the analysis rather than feed it: Requirement E, which most benchmarks fail for want of an exchangeable reference set, and the placebo gate of \S\ref{sec:phase0}, which fails whenever surface separability grows with depth in the model under audit. Drawing the placebo as a gate rather than a control is the change \S\ref{sec:phase0} forced.}
\label{fig:protocol}
\end{figure*}

Figure~\ref{fig:protocol} shows the whole procedure before the pieces are specified individually.

\subsection{Setup and notation}

Let $M$ be a decoder-only transformer with $L$ blocks and hidden width $d$. For a token sequence $x=(x_1,\dots,x_T)$, write $h^{(\ell)}_t(x)\in\Real^d$ for the residual stream at position $t$ after block $\ell$, with $h^{(0)}$ the embedding plus positional contribution and no attention or MLP applied.

A benchmark item is a pair $(q,a)$ where $q$ is everything the model is shown before it must produce an answer, including any few-shot context, instructions, and answer options, and $a$ is the gold continuation. Write $p=q$ for the prefix and $f=q\oplus a$ for the full item.

We are given a \emph{suspect set} $S=\{s_1,\dots,s_n\}$ of items drawn from the benchmark under audit and a \emph{reference set} $R=\{r_1,\dots,r_n\}$ of items believed absent from the model's training data. Assign $y_i=1$ to items of $S$ and $y_i=0$ to items of $R$, giving a pooled sample $\mathcal{D}=\{(x_i,y_i)\}_{i=1}^{2n}$.

The hypothesis under test is
\[
H_0:\ \text{$S$ and $R$ are exchangeable}
\]
given the nuisance features of \S\ref{sec:nuisance}, against the alternative that the model's internal representation separates them beyond what the nuisance features permit. Rejecting $H_0$ is evidence that $M$ represents $S$ differently for reasons not reducible to how $S$ and $R$ differ as text.

\subsection{The exchangeability requirement}
\label{sec:exchange}

Everything downstream depends on the construction of $R$, and no statistical machinery repairs a bad $R$. We state the requirement first because it is the binding constraint on the whole method.

\paragraph{Requirement E.} $R$ must be constructed so that, under the assumption of no contamination, an analyst told only the text of an item could not do better than chance at saying whether it came from $S$ or $R$.

Three constructions satisfy Requirement E to different degrees.

\textbf{E1, randomised injection.} Partition a single item pool at random into an injected half and a withheld half before training. Exchangeability holds by randomisation, exactly, and this is the only construction that is exact. It is available only when the analyst controls training, which is why the calibration stage of RSCP is built on it. The public checkpoints of \citet{oren2024proving} provide a pre-existing instance.

\textbf{E2, commissioned twin.} Have a fresh item set written to match the original on the covariates that drive surface separability. GSM1k \cite{zhang2024gsm1k} was commissioned this way against GSM8K \cite{cobbe2021gsm8k}, matched on human solve rate, number of solution steps, and answer magnitude. Exchangeability is approximate and rests on the matching being complete, which is checkable with the nuisance baseline itself.

\textbf{E3, within-corpus held-out split.} Use a corpus whose train and held-out partitions were drawn from the same distribution, as with Pile train versus Pile validation for the Pythia suite \cite{biderman2023pythia,gao2020pile}. Exchangeability is approximate and known to fail for some Pile subsets, since documents are ordered in time and some subdomains drift, which is one of the cases \citet{das2024blind} document. Per-subdomain analysis with the nuisance baseline reported alongside is mandatory here.

Two constructions that do \emph{not} satisfy Requirement E, and which cover most existing membership-inference evaluation sets, are the temporal split, where $R$ postdates the model cutoff and therefore differs in topic and vocabulary, and the cross-source split, where $S$ and $R$ come from different websites, licences, or extraction pipelines.

An important consequence is that RSCP cannot currently be applied to most benchmarks, because most benchmarks have no exchangeable twin. This is a property of the evaluation ecosystem rather than of the method, and it motivates the benchmark-design recommendation in \S\ref{sec:discussion}.

\subsection{Extraction points and the prefix rule}
\label{sec:extraction}

Define two pooling operators on a sequence $x$ of length $T$:
\begin{align}
E^{\ell}_{\mathrm{last}}(x) &= h^{(\ell)}_{T}(x), \\
E^{\ell}_{\mathrm{mean}}(x) &= \tfrac{1}{T}\textstyle\sum_{t=1}^{T} h^{(\ell)}_{t}(x).
\end{align}
Applying each to the prefix $p$ or the full item $f$ gives four candidate extraction points. RSCP requires a prefix-based point, $\pi\in\{(\mathrm{pre},\mathrm{last}),(\mathrm{pre},\mathrm{mean})\}$, and forbids full-item points.

The argument is a conditioning argument. Let $Z\in\{0,1\}$ be membership and $C\in\{0,1\}$ the event that $M$ answers the item correctly. Contamination inflates scores, so $Z$ and $C$ are dependent by construction; that dependence is the harm we are trying to quantify. The state $h^{(\ell)}_{T}(f)$ is computed from a sequence that contains $a$, and for $\ell$ close to $L$ it is by definition a sufficient statistic for the model's next-token distribution given $q\oplus a$. A probe on that state can therefore approximate a function of the model's scoring of the gold answer, which is a proxy for $C$. Two failures follow.

The first is a confound. If $S$ and $R$ differ in difficulty at all, even slightly, a probe reading a proxy for $C$ separates them without any memorisation being present, and reports it as contamination.

The second is worse and survives perfect difficulty matching. Suppose $S$ and $R$ are matched on intrinsic difficulty but $M$ is contaminated on $S$, so $\Pr(C\mid Z{=}1)>\Pr(C\mid Z{=}0)$. A full-item probe now separates the sets, but it does so by detecting the score inflation itself. The statistic is then a re-description of the accuracy gap, and cannot be used to explain it. The value of a representational detector is that it gives an independent line of evidence about \emph{why} the accuracy gap exists, and that value is destroyed by any extraction point that can see the answer.

This is the same structure as the pre- versus post-continuation artefact identified in the author's earlier layer-wise probing work on Japanese-English pragmatic reasoning \cite{braun2026crosslingual}, where probes read after a generated continuation absorbed information the model had itself produced, inflating apparent encoding of the target property. Here the feedback runs through the answer tokens rather than through a generated continuation, and the fix is the same: extract strictly upstream of any token whose identity or scoring depends on the property under test.

We recommend $(\mathrm{pre},\mathrm{last})$ as the primary point, with $(\mathrm{pre},\mathrm{mean})$ reported as a robustness check, and we recommend reporting the full-item points as well, clearly marked as diagnostics of the confound rather than as contamination estimates. The gap between the prefix and full-item statistics is itself informative: a large gap is a sign that whatever the full-item probe found is competence rather than familiarity.

\subsection{The probe}

Fix an extraction point $\pi$ and a layer $\ell$. Let $\phi_\ell(x)\in\Real^d$ be the extracted activation, standardised feature-wise using training-fold statistics only. The probe is ridge regression on the $\pm 1$-coded label,
\begin{equation}
\hat w_\ell = \arg\min_{w\in\Real^d}\ \|X_\ell w - \tilde y\|_2^2 + \lambda\|w\|_2^2 ,
\label{eq:ridge}
\end{equation}
where $X_\ell\in\Real^{2n\times d}$ stacks the standardised activations and $\tilde y = 2y-1$.

We use ridge rather than logistic regression for a reason that becomes important in \S\ref{sec:inference}: the fitted values are linear in the labels. Partition $\mathcal{D}$ into $K$ folds and let $k(i)$ be the fold containing item $i$, let $X^{(-k)}_\ell$ stack the rows outside fold $k$, let $P_{-k}$ be the selection matrix returning those rows to their positions in $\mathcal{D}$, and write
\begin{equation}
A^{(-k)}_\ell = \big(X_{\ell}^{(-k)\top}X_{\ell}^{(-k)}+\lambda I\big)^{-1}.
\end{equation}
The cross-fitted smoother $\Pi_\ell\in\Real^{2n\times 2n}$ has rows
\begin{equation}
(\Pi_\ell)_{i\cdot} = x_{i}^{\top}\,A^{(-k(i))}_\ell\,X_{\ell}^{(-k(i))\top}P_{-k(i)} .
\label{eq:smoother}
\end{equation}
By construction $(\Pi_\ell)_{ij}=0$ whenever $k(i)=k(j)$, so every prediction is out of fold, and $\Pi_\ell$ does not depend on $\tilde y$. That second property is what \S\ref{sec:inference} exploits.

Out-of-fold scores are $\hat u_\ell = \Pi_\ell \tilde y$. Writing $n_1=\sum_i y_i$ and $n_0=2n-n_1$, the probe's balanced accuracy is
\begin{equation}
\BA_\ell(\tilde y) = \frac{1}{2n_1}\sum_{i:y_i=1}\!\mathbf{1}[\hat u_{\ell i}>\tau]
 + \frac{1}{2n_0}\sum_{i:y_i=0}\!\mathbf{1}[\hat u_{\ell i}\le\tau]
\label{eq:ba}
\end{equation}
with $\tau$ the median of $\hat u_\ell$, which keeps the statistic invariant to a constant shift and avoids a threshold chosen on the same data.

The regularisation $\lambda$ is chosen on an inner split of the training folds only. Fold assignment is repeated over five seeds and all reported quantities are means with standard deviations over seeds, following the convention of the author's earlier probing work.

\subsection{The nuisance family and excess separability}
\label{sec:nuisance}

Let $\Phi_0$ be a declared family of model-independent or memorisation-independent feature maps:
\begin{itemize}\itemsep1pt
\item $\psi_{\mathrm{bag}}$: mean-pooled layer-0 activations over the prefix. This is a bag of token embeddings with no attention and no MLP, so it carries no information the model computed. It is the natural floor for a representational claim.
\item $\psi_{\mathrm{char}}$: $\ell_2$-normalised character 3- to 5-gram TF-IDF.
\item $\psi_{\mathrm{surf}}$: token count, character count, digit ratio, punctuation ratio, type-token ratio, mean word length.
\item $\psi_{\mathrm{ref}}$: per-token log-likelihood summaries (mean, variance, minimum-$k$\% mean for $k\in\{10,20\}$, zlib-normalised ratio) under an independent reference model $M_{\mathrm{ref}}$ that has not been trained on any candidate corpus.
\end{itemize}

The nuisance baseline is the cross-fitted balanced accuracy of the same ridge probe on the concatenation of all four, using the same folds and the same seeds:
\begin{equation}
\BA_{\nuis} = \BA\big(\text{concat}(\psi_{\mathrm{bag}},\psi_{\mathrm{char}},\psi_{\mathrm{surf}},\psi_{\mathrm{ref}})\big).
\end{equation}
Excess separability at layer $\ell$ is
\begin{equation}
\Delta_\ell = \BA_\ell - \BA_{\nuis} .
\label{eq:delta}
\end{equation}
$\Delta_\ell$ is reported, but it is not tested. \S\ref{sec:phase0} measures the obvious statistic built on it, $\max_\ell \Delta_\ell$, and finds it unusable: $\BA_{\nuis}$ depends on the dimension of the nuisance block, so the level of $\Delta_\ell$ is biased in a direction set by an implementation detail. What is tested instead is a zero-sum contrast on the depth profile, recentred on a level-matched placebo baseline, given in Eq.~\eqref{eq:adj}.

Two points about $\psi_{\mathrm{ref}}$. It is deliberately a strong adversary: it includes the statistics that likelihood-based membership inference uses \cite{shi2024minkprob}, computed on a model that cannot have memorised the items. Whatever separability those statistics achieve on $M_{\mathrm{ref}}$ is attributable to how the two sets differ as text. Setting the bar there is conservative, and it is the point of the construction. It also inherits a weakness: if $M_{\mathrm{ref}}$ is itself contaminated on $S$, $\BA_{\nuis}$ is inflated and $\Delta_\ell$ is biased downward, making the test conservative rather than anti-conservative. We prefer that direction of error and note it in \S\ref{sec:limitations}.

\paragraph{Required controls.} Three further checks accompany every reported $\Delta$.

\emph{Control task.} Replace $y$ with labels drawn uniformly at random, independent of set membership, and rerun the whole pipeline. Selectivity is $\Delta_{\ell}$ minus the control-task value \cite{hewitt2019control}. A pipeline that produces positive selectivity on random labels has a leak, usually in the standardisation or fold construction.

\emph{Random-direction baseline.} Replace $\phi_\ell$ by its projection onto a random $r$-dimensional subspace for $r\in\{16,64,256\}$ and by projections onto single random directions. This checks that $T$ is not an artefact of $d$ being large relative to $n$, and follows the random-direction convention of the author's earlier probing work.

\emph{Internal null.} Split $R$ at random into halves $R_1,R_2$ and run the entire protocol with $S\!\leftarrow\!R_1$, $R\!\leftarrow\!R_2$. This is an exact instance of E1, so $T$ should be near zero and the permutation $p$-value should be uniform. Twenty replicates with a Kolmogorov-Smirnov test against uniformity is a cheap and effective end-to-end pipeline test.

\emph{Length matching.} Stratify by prefix token length into deciles and subsample to equalise the two sets' length distributions, then report the length-matched $T$.

\subsection{Inference}
\label{sec:inference}

Under $H_0$ the labels are exchangeable, which licenses a permutation test. Draw $B$ permutations $\sigma_b$ of $\{1,\dots,2n\}$, form $\tilde y^{(b)}=\tilde y_{\sigma_b}$, and recompute
\begin{equation}
T^{(b)} = \max_{\ell}\Big[\BA_\ell(\tilde y^{(b)}) - \BA_{\nuis}(\tilde y^{(b)})\Big].
\end{equation}
The $p$-value is
\begin{equation}
\hat p = \frac{1+\#\{b: T^{(b)}\ge T\}}{1+B}.
\end{equation}

The statistic actually tested is the recentred contrast of Eq.~\eqref{eq:adj}, a single pre-registered scalar, so no correction across layers is needed. What the permutation supplies that a bootstrap does not is the variance of the probe fit: $\Pi_\ell$ is label-free, so $\Pi_\ell\tilde y^{(b)}$ is the out-of-fold prediction of a probe refitted to the permuted labels, and \S\ref{sec:phase0} measures the difference this makes. When several benchmarks or models are audited, apply Benjamini-Hochberg across the resulting $p$-values \cite{benjamini1995fdr}.

\paragraph{Cost.} Naively this requires refitting $B\times(L+1)\times K$ probes. Equation~\eqref{eq:smoother} removes that. Because $\Pi_\ell$ is independent of the labels, it is computed once per layer per fold-seed, at cost $O(K(nd^2+d^3))$ or $O(Kn^3)$ via the dual form when $d>n$, and every permutation thereafter costs a single matrix-vector product $\Pi_\ell\tilde y^{(b)}$ at $O(n^2)$, plus a sort for the median threshold. For $n=1000$ per set, $L=24$, $K=5$, $B=1000$, the permutation stage is on the order of $10^{11}$ floating-point operations, which is seconds on a laptop. Activation extraction, done once, dominates.

This is the practical reason for choosing ridge over logistic regression. Logistic fitted values are not linear in the labels, so the shortcut does not apply and the permutation test becomes the expensive part of the protocol. In the linear-probe regime the two objectives generally rank layers similarly, but this is a design choice rather than an established fact and \S\ref{sec:programme} lists it as something to check.

\subsection{The baseline is an estimate, and its variance has to be carried}
\label{sec:basevar}

The permutation above holds the placebo baseline fixed. It is not fixed. The baseline comes from a split of the reference set at the extremes of a surface key, at a noise level chosen by a search, and both the split and the search are random. Its sampling variability therefore belongs in the null and is absent from it.

Whether that matters is an empirical question with an uncomfortable answer. Re-estimating the baseline over $J$ split seeds and taking the standard deviation of $w^{\top}\hat b_j$ gives, on the real audits of \S\ref{sec:phase3}, a quantity \PthreeRatioLo{} to \PthreeRatioHi{} times the standard deviation of the permutation null itself, in every arm measured. The correction is more variable than the thing it corrects. \S\ref{sec:phase0} did not see this because its simulated surface key is derived from the covariate that actually drives item variation; on real text the key is a proxy, and a poor proxy makes the split nearly arbitrary. Sweeping key quality in simulation reproduces the effect, raising the baseline's standard deviation about twentyfold, while sweeping depth at a good key does not.

The protocol therefore requires two additions.

\textbf{Report the ratio.} Measure $\hat\sigma_{b}$ over at least eight split seeds and report it beside the null's own $\hat\sigma_{0}$. A ratio near zero means the recentring is essentially free. A ratio near or above one means the reported $p$ is not the uncertainty of interest.

\textbf{Propagate it.} Widen the null by the measured baseline variance,
\begin{equation}
\tilde T^{(b)} = T^{(b)} + \varepsilon_b,\qquad \varepsilon_b\sim\mathcal{N}(0,\hat\sigma_{b}^{2}),
\end{equation}
and take the $p$-value from $\tilde T$. This assumes the baseline's error is independent of the permutation null and approximately Gaussian. Both are approximations; treating an estimated quantity as known is not an approximation but an error, and \S\ref{sec:phase3} shows it is the difference between a significant finding and none.

A verdict that survives only while $\hat b$ is treated as known is not admissible, and the released implementation refuses to issue one, in the same way it refuses when Requirement E fails.

\subsection{Exposure calibration}
\label{sec:calibration}

A binary contaminated/clean verdict is not the right output, because contamination is graded and because a null result at one exposure level says nothing about another. RSCP therefore reports an interval on effective exposure.

On a family of calibration models trained with known duplication counts $m\in\{0,1,2,4,8,16,32\}$ (construction E1), fit a monotone saturating link
\begin{equation}
g(m;\gamma,\beta) = \gamma\big(1-e^{-\beta m}\big),\qquad g(0)=0,
\label{eq:link}
\end{equation}
to the observed $(m, T)$ pairs, with $\gamma$ the asymptotic excess separability and $\beta$ the exposure sensitivity. The saturating form encodes the expectation that additional repetitions have diminishing effect once an item is memorised, consistent with the duplication-count dependence of memorisation reported by \citet{carlini2023quantifying} and \citet{kandpal2022dedup}.

For a suspect benchmark with observed $T$ and permutation-derived null spread, the reported quantity is the set
\begin{equation}
\widehat{\mathcal{M}} = \{m\ :\ T \in \mathrm{CI}_{95}\big(g(m;\hat\gamma,\hat\beta)\big)\},
\end{equation}
read as ``the observed excess separability is consistent with an effective exposure of $\widehat{\mathcal{M}}$ repetitions''. Two quantities from the calibration are worth reporting on their own. The \emph{detection floor} $m^{*}$ is the smallest $m$ at which the test rejects at $\alpha=0.05$ with power $0.8$ at the given $n$; it states what the instrument can and cannot see. The \emph{saturation point} is the $m$ beyond which $g$ is flat, beyond which the method cannot distinguish heavier contamination.

Transfer of $\hat\gamma,\hat\beta$ from calibration models to a target model is an assumption, not a result, and is the weakest link in the procedure. We recommend calibrating within model family and scale wherever possible, and reporting $\widehat{\mathcal{M}}$ only when a same-family calibration exists.

\subsection{Item margins and contamination-adjusted scores}
\label{sec:adjusted}

The out-of-fold score $\hat u_{\ell^{*}i}$ gives a per-item margin. Individually these are noisy, for the reasons \citet{duan2024mia} document, and we do not propose per-item verdicts. Ranked, they support a set-level comparison.

Let $S_{\mathrm{hi}}$ and $S_{\mathrm{lo}}$ be the top and bottom margin quartiles of $S$. The raw inflation estimate is
\begin{equation}
\widehat{\mathrm{Infl}} = \mathrm{acc}(S_{\mathrm{hi}}) - \mathrm{acc}(S_{\mathrm{lo}}).
\end{equation}
This is circular if margin correlates with item difficulty, which it plausibly does, since easy items are shorter, more templated, and more likely to recur online. The control is difficulty matching against a panel. Estimate a difficulty score $b_i$ for each item as the fraction of a panel of reference models, chosen to predate the benchmark or to have verified clean corpora, that answer it correctly, then resample $S_{\mathrm{hi}}$ and $S_{\mathrm{lo}}$ to equalise the distribution of $b$ and recompute. Report both, and report the panel. The difficulty estimate can equivalently be an item-response-theory difficulty parameter fitted across the panel, which is the natural point of contact between this method and psychometric approaches to benchmark analysis \cite{polo2024tinybenchmarks}.

Regression to the mean will produce a positive raw $\widehat{\mathrm{Infl}}$ even with no contamination whenever accuracy is measured on the same items used to fit the margin, so the accuracy evaluation must be done on a held-out fold, using the same fold structure as the probe.

\subsection{The protocol as a checklist}

Table~\ref{tab:checklist} states the procedure compactly. Steps 1, 3 and 7 are the ones most often skipped, and a result reported without them should be read as an upper bound on the memorisation signal rather than an estimate of it.

\begin{table}[t]
\small
\begin{tabular}{@{}p{0.05\columnwidth}p{0.87\columnwidth}@{}}
\toprule
1 & Construct $R$ satisfying Requirement E, with $|R| = 2|S|$. \\
2 & Extract prefix activations at all layers, $(\mathrm{pre},\mathrm{last})$ primary. \\
3 & Fit the nuisance probe; report $\BA_{\nuis}$ before anything else. \\
4 & Fit cross-fitted ridge probes at each layer over 5 fold-seeds. \\
5 & Report $\Delta_\ell$ and the profile $\BA_\ell$ descriptively. \\
6 & Build the level-matched placebo baseline on the other half of $R$. \\
7 & Test the recentred contrast (Eq.~\ref{eq:adj}) against the permutation null. \\
8 & Run the four controls: control task, random direction, internal null, length matching. \\
9 & If a same-family calibration exists, report $\widehat{\mathcal{M}}$ and $m^{*}$. \\
10 & Report the full-item extraction points as confound diagnostics. \\
11 & If reporting adjusted scores, report both raw and difficulty-matched inflation, with the panel. \\
\bottomrule
\end{tabular}
\caption{The RSCP protocol. Steps 1, 3, 6 and 7 are the ones \S\ref{sec:phase0} shows cannot be skipped.}
\label{tab:checklist}
\end{table}

\section{Validation, and the Four Choices It Forced}
\label{sec:phase0}

Everything above the line is a proposal. One part of it can be settled without
a GPU: whether the statistical procedure does what it claims. This section
reports that, and it is the reason the protocol looks the way it does. Each of
the four design choices in \S\ref{sec:method} was adopted because a simpler
alternative was measured and failed.

Measurements of a statistical procedure are only as good as the range of data
they were taken over, so everything is run under two generative models. Sim-A
is deliberately simple: isotropic Gaussian layers, one nuisance direction,
layers conditionally independent. Sim-B is a residual stream with the
pathologies real transformers have. Table~\ref{tab:sims} shows they are in
different regimes, which is a measurement rather than an assertion. Where a
conclusion holds under both we report it as a property of the procedure; where
it holds under one only, we say so.

\begin{table}[t]
\small
\centering
\begin{tabular}{@{}lcc@{}}
\toprule
& Sim-A & Sim-B \\
\midrule
residual-stream norm growth & $\times\PropAGrowth$ & $\times\PropBGrowth$ \\
correlation, layer 1 to $L$ & \PropACorr & \PropBCorr \\
kurtosis of activations & \PropAKurt & \PropBKurt \\
share of variance, top eigenvector & \PropAEig & \PropBEig \\
\bottomrule
\end{tabular}
\caption{The two simulators occupy different regimes. Sim-B adds anisotropic
power-law covariance, heavy-tailed innovations, growing per-layer scale,
several correlated nuisance directions, and accumulation down the residual
stream.}
\label{tab:sims}
\end{table}

In both, the nuisance is a continuous item covariate whose distribution
differs between the two sets, not a property of membership. That detail is
load-bearing. If surface variation were tied to the label, depth-dependent
surface separability would exist only along the suspect-reference axis, no
within-reference split could ever observe it, and a placebo baseline would be
useless by construction. An earlier version of these simulators made exactly
that mistake and produced a misleading negative result.

Throughout: $d=96$, $L=12$, $\NPerSet$ suspect items, twice that many
reference items, $\BPerm$ permutations, $\alpha=0.05$, and 40 to 60 replicates
per cell. The whole study is about \RunMinutes{} minutes on four CPU cores.

\subsection{C1: the level of separability is unusable}

\input{generated/results_tables.tex}

The obvious statistic is excess separability, $\Delta_\ell = \BA_\ell -
\BA_{\nuis}$, aggregated over layers. It does not work, and the reason has
nothing to do with contamination. At matched effective degrees of freedom a
lower-dimensional feature block still estimates its discriminative direction
from a smaller space and reaches higher accuracy at equal population
separation, so $\BA_{\nuis}$ depends on how large the analyst made the control
set, and $\Delta_\ell$ inherits that as a bias.

\begin{figure}[t]
\centering
\includegraphics[width=\columnwidth]{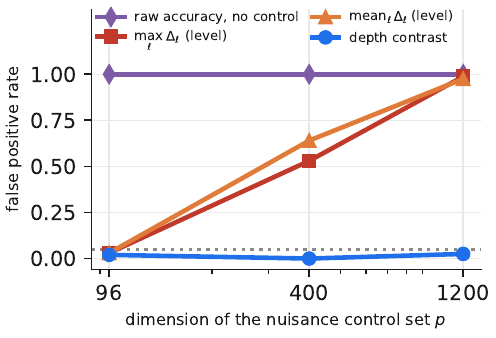}
\caption{\textbf{C1.} False positive rate under a true null as the nuisance
control set grows. The zero-sum contrast is flat because the control term
cancels algebraically; every level-based statistic tracks the control set's
dimension.}
\label{fig:c1}
\end{figure}

Table~\ref{tab:c1} and Figure~\ref{fig:c1} measure it. Under a true null the
max-level statistic rejects \CapMaxMatch{} of the time against a
dimension-matched control and \CapMaxHuge{} against a control of
$p=\CapPdimHi$. The direction is what matters: the nuisance family we
recommend contains a character $n$-gram block far larger than the activations,
so a protocol reporting the level would sit at the right-hand end of that
figure and manufacture contamination on clean models. Subtracting $\BA_{\nuis}$
does not repair the naive statistic, which rejects \CapNaiveMatch{} everywhere;
it converts a guaranteed error into one that depends on an arbitrary
implementation choice, which is worse because it looks principled.

A contrast with weights summing to zero removes the problem exactly, since
$\sum_\ell w_\ell (\BA_\ell - \BA_{\nuis}) = \sum_\ell w_\ell \BA_\ell$. It
also changes what is being tested, from the level of separability to its shape
in depth, which is the quantity that distinguishes memorisation from surface
difference in the first place.

\subsection{C2: depth profiles are not flat, and fail both ways}

A contrast against a fixed rising ramp assumes surface decodability is flat in
depth. That assumption fails, and it fails in both directions.

Make surface separability grow with depth and the uncorrected contrast rejects
a true null \SlopeARaw{} of the time at 25 per cent growth and \SlopeARawFifty{}
at 50 per cent. Run instead on a residual stream where the surface signal is
written once at the embedding and noise accumulates on top of it, and the
profile declines: under Sim-B a true null falls from \ProfNullFirstB{} at the
embedding to \ProfNullLastB{} at the final layer (Figure~\ref{fig:c4}). A
declining baseline makes the contrast negative before any memorisation is
present, and the test loses all of its power rather than gaining false
positives.

The fix does not add an assumption, it removes one. The placebo split already
estimates the baseline: two halves of the reference set, every item a
non-member, so whatever depth profile appears there is the profile of surface
separability in this model. Recentring on it,
\begin{equation}
T_{\mathrm{adj}} \;=\; \sum_{\ell} w_\ell\big(\BA_\ell - \BA^{\mathrm{plc}}_\ell\big),
\label{eq:adj}
\end{equation}
replaces an untestable global claim with a locally estimated correction. The
placebo stops being a veto and becomes the baseline.

\begin{figure}[t]
\centering
\includegraphics[width=\columnwidth]{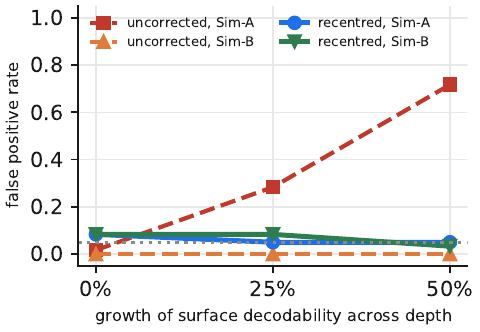}
\caption{\textbf{C2.} Recentring holds the nominal rate across the whole range
where the uncorrected contrast fails. Measured under both simulators, true
null throughout.}
\label{fig:c2}
\end{figure}

Table~\ref{tab:c2} and Figure~\ref{fig:c2} give the result. Recentring holds
the false positive rate at \SlopeAAdjFifty{} where the uncorrected statistic
reaches \SlopeARawFifty{}, and it does so without costing power: Sim-B, where
the uncorrected contrast had none, reaches \PwBFifteen{} at $\varepsilon=1.5$.

\subsection{C3: the null must refit the probe}

The natural null for a statistic computed on items is a bootstrap over items.
It is wrong here. Resampling items with the fitted probe held fixed captures
evaluation variance but not the variance of the fit, and the omission is not
small: under a true null the item bootstrap rejects between \BootMin{} and
\BootMax{} of the time across the same cells where the permutation null holds
\CapContrastMatch{} to \CapContrastHuge{}.

Permuting labels against the cross-fitted smoother captures both, because
$\Pi_\ell$ is label-free and $\Pi_\ell \tilde y^{(b)}$ is therefore the
out-of-fold prediction of a probe \emph{refitted} to the permuted labels. The
same property that makes the permutation cheap is what makes it correct. The
measured rate is \CapContrastMatch{} to \CapContrastHuge{} across the range in
Figure~\ref{fig:c1}.

\subsection{C4: the baseline needs the same sample size}

A baseline estimated on half as many items as the comparison is not
comparable, because probe accuracy depends on sample size and the dependence
differs across the profile. Splitting the reference set in half to build the
placebo, while comparing at full size, gives a false positive rate of
\SizeHalfA{} under Sim-A against \SizeMatchedA{} when the sizes match. The
protocol therefore requires $|R| = 2|S|$: half the reference set is the
comparison arm, half is the baseline arm. Under Sim-B the two agreed, so this
binds in some regimes and not others, and we impose it because it is free at
design time.

Matching size is necessary and not sufficient. Splitting the reference set at
the extremes of a surface key separates the halves more strongly than the two
sets actually differ, and a stronger separation bends the profile differently.
We therefore search over how coarsely to split and take the split whose
embedding-layer separability is closest to the observed value. Layer 0 is the
right anchor because it is a bag of token embeddings and carries surface
information only.

\subsection{What the instrument can see}

\begin{figure}[t]
\centering
\includegraphics[width=\columnwidth]{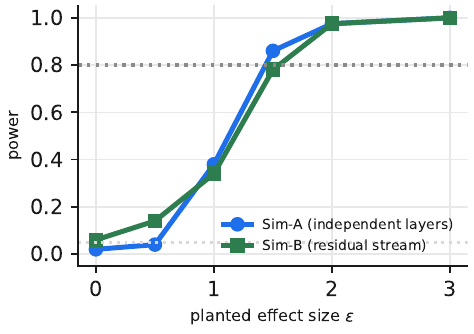}
\caption{Power of the recentred contrast under both simulators. Both reach
0.8 near $\varepsilon = 1.5$ and saturate by $\varepsilon = 3$.}
\label{fig:c3}
\end{figure}

\begin{figure*}[t]
\centering
\includegraphics[width=0.94\textwidth]{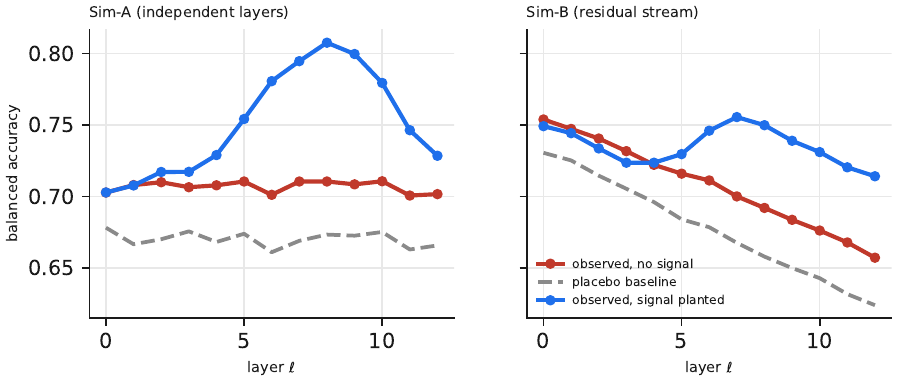}
\caption{Recovered depth profiles. Left, independent layers: the null profile
is flat and the placebo tracks it. Right, a residual stream: the null profile
declines by roughly nine accuracy points from embedding to final layer, which
is the failure mode C2 corrects. In both, the planted signal is recovered with
its peak near the planted layer.}
\label{fig:c4}
\end{figure*}

Figure~\ref{fig:c3} gives power against planted effect size. Both simulators
cross 0.8 near $\varepsilon = 1.5$ (\PwAFifteen{} and \PwBFifteen{}) and reach
\PwAThirty{} by $\varepsilon = 3$. Below $\varepsilon \approx 1$ the instrument
is underpowered, which is the resolution it has and the reason a null result
has to be reported with a detection floor beside it.

Figure~\ref{fig:c4} shows the profiles the contrast weights, and makes the C2
problem visible directly: the two panels have null profiles of opposite slope,
and only a baseline estimated per model can accommodate both.

\subsection{The placebo is only as good as the surface key}

\begin{figure}[t]
\centering
\includegraphics[width=\columnwidth]{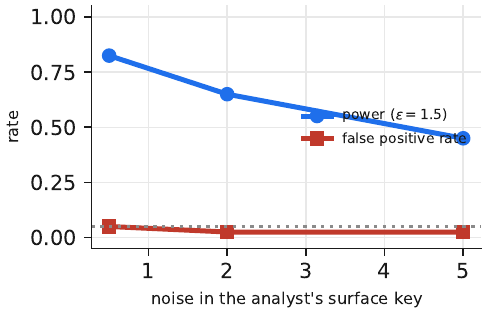}
\caption{Degrading the analyst's surface key costs power and does not cost
validity. Measured under Sim-B.}
\label{fig:c5}
\end{figure}

The baseline is built by splitting the reference set on a variable the analyst
chooses, and that choice is not neutral. Degrading the key from a close
correlate of the underlying covariate to a loose one takes power from
\KeyPwFive{} to \KeyPwFifty{} while the false positive rate stays at or below
\KeyFprFive{} (Figure~\ref{fig:c5}). The failure mode is therefore benign: a
poor key makes the instrument blind, not wrong. This is worth stating plainly
because the opposite would make the method unusable in practice, where the
analyst rarely knows which surface feature separates the two sets.

\subsection{Cost}

At $n=\TimeN$ items per set and $d=\TimeD$, building the cross-fitted smoother
takes \TimeSetup~s, after which each permutation costs \TimePerPerm~ms and
$2000$ of them cost \TimeB~s. Refitting per permutation, which any probe
without label-linear fitted values forces, projects to \TimeRefitH~hours, a
factor of about \TimeSpeedup. That is the difference between running the
control battery and skipping it, and it is why the battery is mandatory here
rather than encouraged.

\subsection{Honest residuals}

Three things are not fixed. Calibration is close to nominal but not exact,
sitting between \SlopeAAdjZero{} and \SlopeBAdjZero{} across the null cells,
so $p$-values near the threshold should not be read as exact. Power under a
realistic residual stream requires a larger effect than under independent
layers, and we have no way to convert either $\varepsilon$ scale into a
duplication count without the injection experiments of \S\ref{sec:programme}.
And every number here is a measurement of a procedure on synthetic
activations. None of it is evidence that a real transformer carries a
familiarity direction at all.

\section{Worked Examples}
\label{sec:worked}

Three settings are described. The first calibrates the instrument against a corpus whose contents are known. The second uses existing public checkpoints with deliberately injected contamination at known duplication counts. The third applies the protocol to a benchmark of live interest.

Numbers attributed to published work are real measurements and are cited. Numbers in tables marked \emph{illustrative} are synthetic. They are there to show the shape of the output and to make the reporting format concrete; the study in \S\ref{sec:programme} would produce the real values.

\subsection{WE1: Pythia and the Pile}
\label{sec:we1}

\paragraph{Why this setting.} The Pythia suite \cite{biderman2023pythia} trains sixteen models from 70M to 12B parameters on the Pile \cite{gao2020pile} in a fixed data order, releases 154 checkpoints per model, and releases tooling to reconstruct the exact training dataloader. Three properties follow that no other public suite offers together. Membership is known exactly. The step at which a sequence was seen is known, which allows exposure recency to be varied. And the deduplicated variants of each model give a second, independent handle on duplication count, since deduplication removes exactly the repeated exposures that drive memorisation \cite{lee2022dedup}. Which sequences a Pythia model memorises is partly predictable from smaller models in the same suite \cite{biderman2023emergent}, which gives an additional external quantity for $T$ to be validated against. Seed-level variability in this suite has been characterised directly \cite{vanderwal2025polypythias}, so the fold-seed variance we report can be compared against the pretraining-seed variance that bounds it.

\paragraph{Design.} Take $S$ from the Pile training split and $R$ from the Pile validation split, matched within subdomain, so construction E3 applies. Run per subdomain across the drift spectrum, PubMed Central and Wikipedia at the stable end, GitHub and arXiv at the drifting end, and report $\BA_{\nuis}$ per subdomain first. Extract at $(\mathrm{pre},\mathrm{last})$ for Pythia-410M, 1.4B and 2.8B and their deduplicated counterparts at every layer. Storage is modest: 2000 items times 25 layers times 2048 dimensions in half precision is about 200~MB per model.

\paragraph{Expected outcome.} \citet{duan2024mia} evaluate this exact setting for likelihood-based membership inference across the Pythia suite and find performance barely above chance, which they attribute to the very large corpus, close to a single pass over it, and a fuzzy member boundary. The honest prior for WE1 is therefore that $T$ is small. That is not a failure of the experiment. WE1's purpose is to establish the noise floor and to confirm that the internal-null check produces uniform $p$-values on a real corpus, which is the precondition for believing anything the protocol says in WE2 and WE3.

\paragraph{Recency sub-experiment.} Because the training step at which each sequence was seen is known, $S$ can be stratified by exposure step and $T$ computed per stratum, giving a forgetting curve for representational familiarity. This connects to the causal treatment of memorisation profiles over Pythia's data order given by \citet{lesci2024causal} and to the forgetting dynamics of \citet{jagielski2023forgetting}. A monotone decay of $T$ with time since exposure would be strong evidence that $T$ is tracking memorisation rather than a static property of the text, and is the single most convincing validity check available in this setting.

\subsection{WE2: deliberately contaminated checkpoints}
\label{sec:we2}

\paragraph{Why this setting.} \citet{oren2024proving} released models pretrained with intentional injection of benchmark data, at 1.4B and 124M parameters, together with the injected datasets and their duplication counts. Because injected and clean items come from the same benchmarks and the assignment is not temporal, this is an instance of E1 and exchangeability holds without further work. As reported by \citet{liu2024probing}, PubMedQA \cite{jin2019pubmedqa} was injected once and CommonsenseQA \cite{talmor2019csqa} twice.

\paragraph{Published reference point.} Table~\ref{tab:we2published} reproduces the AUCs \citet{liu2024probing} measured on these checkpoints. Every method, including their activation probe, sits within a few points of chance. This is the most informative published result on what activation probing can do under an exchangeable split at realistic duplication counts, and it should anchor expectations at $m\in\{1,2\}$.

\begin{table}[t]
\small
\begin{tabular}{@{}lcc@{}}
\toprule
Method & PubMedQA & CommonsenseQA \\
       & ($m{=}1$) & ($m{=}2$) \\
\midrule
Loss attack            & 48.0 & 49.9 \\
Neighbour attack       & 53.0 & 50.0 \\
Min-$k$\% Prob         & 47.5 & 49.6 \\
zlib ratio             & 46.1 & 48.8 \\
Lowercase ratio        & 50.7 & 49.2 \\
Smaller-model ratio    & 49.5 & 49.5 \\
Activation probe       & 54.0 & 51.9 \\
\bottomrule
\end{tabular}
\caption{Published AUC (\%) on the deliberately contaminated checkpoints of \citet{oren2024proving}, as measured by \citet{liu2024probing}, Table 4. Chance is 50.0. These are real measurements, not illustrative. $m$ is the injection duplication count.}
\label{tab:we2published}
\end{table}

\paragraph{Design.} Apply RSCP with $S$ the injected items and $R$ the withheld items of the same benchmark. Report $\Delta_\ell$ across all layers of the 1.4B model, with the 124M model as a scale contrast. Two extensions matter.

First, extend the exposure range. The public checkpoints give only $m\in\{1,2\}$, which is exactly the regime where everything is near chance and where the calibration curve is least identifiable. Continued pretraining of Pythia-160M and Pythia-410M on a held-out slice of the Pile with a benchmark injected at $m\in\{0,1,2,4,8,16,32\}$ fills in the curve. Randomised assignment of items to injected and withheld halves gives exact exchangeability.

Second, vary the injection format, since contamination in the wild does not arrive as clean question-answer pairs. Four conditions: verbatim item text; item text in evaluation format with the gold answer; a paraphrase from an independent model, following \citet{yang2023rephrased}; and the source document without question framing. The paraphrase condition is what distinguishes RSCP from overlap search, which is blind to it by construction.

\begin{table}[t]
\small
\setlength{\tabcolsep}{4pt}
\begin{tabular}{@{}lcccc@{}}
\toprule
$m$ & Verbatim & Eval-fmt & Paraphrase & Source doc \\
\midrule
0  & 0.00 & 0.00 & 0.00 & 0.00 \\
1  & 0.01 & 0.02 & 0.00 & 0.00 \\
2  & 0.02 & 0.03 & 0.01 & 0.01 \\
4  & 0.05 & 0.07 & 0.02 & 0.01 \\
8  & 0.11 & 0.14 & 0.05 & 0.02 \\
16 & 0.18 & 0.22 & 0.09 & 0.04 \\
32 & 0.23 & 0.27 & 0.13 & 0.06 \\
\bottomrule
\end{tabular}
\caption{\illus{} Shape of the exposure-response table for WE2 ($T$, Pythia-410M continued pretraining). Values are synthetic and are shown only to define the reporting format and to make the fitted link in Eq.~\eqref{eq:link} concrete. The quantities of scientific interest are the detection floor $m^{*}$ in each column and the ordering of the four columns, both of which the study in \S\ref{sec:programme} would determine.}
\label{tab:we2}
\end{table}

The deliverable of WE2 is the detection floor $m^{*}$ per condition, the number that tells practitioners when an activation probe is the wrong tool.

\subsection{WE3: GSM8K against GSM1k}
\label{sec:we3}

\paragraph{Why this setting.} GSM1k \cite{zhang2024gsm1k} is a set of 1000 grade-school arithmetic problems commissioned to mirror GSM8K \cite{cobbe2021gsm8k}, matched on human solve rate, number of solution steps, and answer magnitude. It is the clearest existing instance of construction E2. Its authors report accuracy drops of up to 8 points on GSM1k relative to GSM8K, with several model families showing systematic gaps across sizes, and a positive relationship (Spearman $r^2=0.36$) between a model's probability of generating a GSM8K example and its GSM8K-to-GSM1k gap. Frontier families show little or no gap.

\paragraph{What RSCP adds.} The GSM1k evidence is behavioural. A gap on a matched fresh set is consistent with memorisation, but equally with imperfect matching, with GSM1k being harder along an unmatched dimension, or with overfitting to GSM8K's surface conventions. These have different consequences for how to read a GSM8K number, and behavioural evidence cannot separate them.

RSCP gives a second, representational line of evidence on the same split, and the two lines can disagree in an informative way:

\begin{figure*}[t]
\centering
\includegraphics[width=0.98\textwidth]{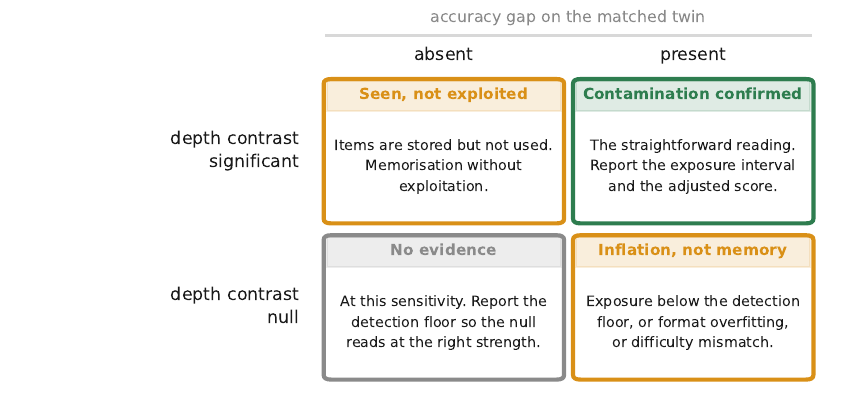}
\caption{The four outcomes and what each licenses. Behavioural evidence alone cannot separate the right-hand column, which is what the representational test adds.}
\label{fig:decision}
\end{figure*}

Figure~\ref{fig:decision} sets out the four outcomes and what each licenses.

\paragraph{Model selection.} Three open models runnable on one consumer GPU: an OLMo model, because Dolma is public and any verdict can be checked against corpus search \cite{groeneveld2024olmo,soldaini2024dolma,elazar2024wimbd}; a Pythia model of comparable scale as a low-contamination control, since the Pile predates GSM8K's wide circulation; and one recent small instruction-tuned model, where the prefix rule applies inside the chat template and extraction is at the last token before the assistant turn.

\paragraph{A check that costs nothing.} Before any probing, compute $\BA_{\nuis}$ for GSM8K against GSM1k. If it is near 0.5 the commissioned matching worked, which is an independent audit of GSM1k that its own construction checks do not provide. If it is well above 0.5, a blind classifier can tell the two sets apart, every behavioural comparison between them carries a caveat, and the responsible features can be read off the fitted nuisance weights. Either result is worth reporting.

\section{Discussion}
\label{sec:discussion}

\subsection{What the protocol says about the benchmarks it is applied to}

\paragraph{On GSM8K.} The field currently treats the GSM8K-to-GSM1k gap as the strongest available evidence that GSM8K is contaminated for some model families. The gap is real and carefully measured. What it does not do is distinguish memorisation of items from adaptation to a format. RSCP makes the distinction operational, and the negative branch is as useful as the positive one: a large gap with $T$ near zero, on a model where the calibration curve says $m^{*}$ is small, would mean the gap is not item memorisation, which would redirect attention from data hygiene to the more awkward question of whether heavy tuning on GSM8K-style data is itself a form of benchmark-specific overfitting that no amount of decontamination fixes.

\paragraph{On MMLU-style benchmarks.} MMLU \cite{hendrycks2021mmlu} was assembled from many independent online sources across 57 subjects, so its items have heterogeneous provenance and there is no reason to expect uniform contamination. The natural unit of analysis is the subject, not the benchmark, and RSCP is applied per subject with Benjamini-Hochberg across subjects. The absence of an exchangeable reference set is the blocker: a fresh matched set would have to be commissioned per subject. Until then, MMLU can be audited only in the E1 direction, by injecting a randomly chosen half into a small model and measuring what exposure level would be needed to produce the separability seen in the wild.

\paragraph{On the Pile-derived settings.} The most likely outcome of WE1 is a set of small, mostly non-significant $T$ values with large $\BA_{\nuis}$ on the drifting subdomains. Reported as excess separability rather than as raw accuracy, this is a clean replication of \citet{duan2024mia} through a different instrument, and it puts a number on how much of the apparent separability in this literature is corpus drift.

\subsection{The most informative negative result}

If the calibration curve of \S\ref{sec:calibration} places $m^{*}$ above the duplication counts that occur in practice, the correct conclusion is that activation probing does not solve low-duplication benchmark contamination, and the paper reporting that should say so plainly. The published evidence points that way: likelihood-based attacks under exchangeable splits perform near chance \cite{duan2024mia}, and the one published activation probe evaluated on a non-temporal split reached 54.0 and 51.9 AUC at $m=1$ and $m=2$ \cite{liu2024probing}. A well-instrumented negative result with a stated detection floor is more useful to the field than another detector reporting high AUC on a split that a blind classifier can also solve.

\subsection{Implications for benchmark design}

Three changes would make contamination auditing tractable, and all are cheap at construction time relative to their value later.

\emph{Publish a matched twin at release.} Reserve a randomly chosen fraction of items at construction and release them separately, or on request under an agreement not to redistribute. This converts every future audit from construction E2, which requires commissioning new items and defending the matching, to construction E1, where exchangeability holds by randomisation. GSM1k \cite{zhang2024gsm1k} shows the value of the twin and also shows the cost of producing one after the fact.

\emph{Publish a canary.} BIG-bench \cite{srivastava2023bigbench} includes a canary string for the purpose and the practice has not spread. None of ARC \cite{clark2018arc}, HellaSwag \cite{zellers2019hellaswag}, MATH \cite{hendrycks2021math} or MMLU carries one, and all four feed aggregate leaderboards \cite{bommasani2023helm}, so contamination findings on them have to be argued indirectly.

\emph{Publish the matching statistics.} With construction documented in enough detail, a $\BA_{\nuis}$ audit can be run by anyone against any candidate reference set.

\subsection{Relation to sample-efficient evaluation}

Item margins from \S\ref{sec:adjusted} produce a per-item score that is not a difficulty estimate and not a discrimination estimate, so it is orthogonal to the psychometric quantities used to prune benchmarks to an informative core \cite{polo2024tinybenchmarks}. A pruned benchmark selected purely for information content could be systematically more or less contaminated than the full set, since the item properties that make an item informative are correlated with the properties that make it likely to have been scraped. Auditing pruned benchmarks for this is a small, tractable study and we list it in \S\ref{sec:programme}.

\section{Limitations and Threats to Validity}
\label{sec:limitations}

\paragraph{L1. Applicability is bounded by the reference set.} RSCP cannot be run on a benchmark with no exchangeable twin, which today is most of them. Every refinement in \S\ref{sec:method} is downstream of a precondition the evaluation ecosystem mostly does not satisfy.

\paragraph{L2. White-box only.} Activations are required, so closed models are out of scope; the black-box test of \citet{oren2024proving} remains the right tool there, with its own restriction to file-level contamination.

\paragraph{L3. A null result is weak evidence of absence.} A linear probe lower-bounds decodability at a chosen extraction point \cite{belinkov2022probing,pimentel2020info}. $T\approx 0$ means the familiarity signal is not linearly decodable from prefix activations, which is compatible with it being present nonlinearly, at a different extraction point, or in a subspace the standardisation suppresses. The detection floor $m^{*}$ should always be reported alongside a null so the null can be read at the right strength.

\paragraph{L3b. The baseline is only as good as the surface key.} The placebo is built by splitting the reference set on a variable the analyst chooses, and a key that does not track whatever actually differs between the two sets gives a baseline describing the wrong null. \S\ref{sec:phase0} measures the direction of that failure and it is benign, costing power (\KeyPwFive{} down to \KeyPwFifty{} as the key degrades) rather than validity. But a weak key can make the instrument blind without announcing it.

\paragraph{L3c. Calibration is close but not exact.} Across the null cells the recentred contrast rejected between \SlopeAAdjZero{} and \SlopeBAdjZero{} of the time rather than exactly $0.05$. Reported $p$-values near the threshold should not be read as exact, and a headline claim should use a stricter level.

\paragraph{L3d. The simulators are not transformers.} Sim-B has heavy tails, anisotropy, norm growth and residual accumulation, and it is still a generative model we wrote. Real activations may fail in ways neither simulator represents, and only Phase 1 will show that.

\paragraph{L4. The nuisance family is finite.} $\Delta_\ell>0$ says the probe beat the nuisance features we declared. It does not say the probe beat every model-independent feature of the text. The statistic is evidence, not proof, and the declared family should be published so others can attack it by adding features. We consider this preferable to the alternative of an undeclared implicit baseline of chance.

\paragraph{L5. Calibration transfer is assumed, not shown.} Converting $T$ to $\widehat{\mathcal{M}}$ requires that $(\hat\gamma,\hat\beta)$ estimated on a small calibration model apply to the target. Memorisation grows with scale \cite{carlini2023quantifying,tirumala2022memorization}, so a calibration fitted at 410M will not transfer to a 70B model, and probably not cleanly across architectures at the same scale. Report $\widehat{\mathcal{M}}$ only with a same-family calibration; otherwise report $T$ and $\hat p$ alone.

\paragraph{L6. Post-training breaks the clean story.} Instruction tuning shifts representations and contamination can enter at that stage. The prefix rule also interacts with chat templates, where template tokens contribute a large constant component to the activation. Calibration models built with pretraining-stage injection only will understate post-training contamination.

\paragraph{L7. Item-level margins are unreliable.} Per-item membership inference on pretraining data is close to undetectable \cite{duan2024mia}. The adjusted-score procedure in \S\ref{sec:adjusted} uses margins only in aggregate over quartiles, and should not be reinterpreted as a per-item verdict, however tempting the ranking looks.

\paragraph{L8. Circularity in the adjusted score.} Margin plausibly correlates with difficulty. The difficulty-matched control depends on the reference panel being uncontaminated, which is the assumption under audit, applied one level up. The matched estimate is better than the raw one and is not free of the problem.

\paragraph{L9. Adversarial evasion is easy.} A developer wanting to hide contamination can remove linear decodability without removing the benefit, since nullspace projection and related methods delete a linearly encoded property while leaving behaviour largely intact \cite{ravfogel2020inlp,elazar2021amnesic}. RSCP is an instrument for analysing models one is not adversarial towards. It is not an audit that survives a motivated adversary and should not be presented as one.

\paragraph{L10. Small models may not show the phenomenon.} If the familiarity direction emerges only above some scale, results at 160M to 2.8B will be uninformative about the models people actually evaluate. The scale trend across the Pythia suite is the diagnostic, and a flat or non-monotone trend should be reported as failure to establish the effect rather than smoothed over.

\paragraph{L11. Propagating the baseline variance is an approximation.} \S\ref{sec:basevar} widens the null by a measured $\hat\sigma_b$ under independence and normality. Neither is established. The baseline and the observed contrast are computed from overlapping reference items, so independence is optimistic and the widening may be too small; against that, $\hat\sigma_b$ estimated from eight seeds is itself noisy. The honest position is that this replaces a known error with a stated approximation, and that a result sitting near the boundary after propagation should be treated as unresolved rather than as either outcome. Estimating the baseline's contribution by a nested resampling scheme, at proportionate cost, is the obvious improvement and is not done here.

\paragraph{L12. The placebo can only match where the surface key reaches.} The baseline is built by splitting the reference set at the extremes of a surface key, and the search picks whichever split lands closest to the observed layer-0 separability. Both the search and its target are therefore bounded by what that key can produce. Phase 3 shows the failure in both directions: on the full-record arm the match is within \PthreeGapLo{}, but on the short-prefix arm the placebo reaches only $0.497$ against an observed $0.515$, and in an earlier run at half the sample size the observed profile fell below chance, which a key-based split cannot reproduce at all because splitting on any key produces some separability. When the observed level lies outside the key's attainable range the recentring is approximate, and the residual mismatch is a bias of unknown sign. The match gap is reported for exactly this reason and should be read before the contrast.

\paragraph{L13. Multiplicity.} The contrast is a single pre-registered test per model and item pool, but auditing many benchmarks, models and extraction points reintroduces multiplicity; fix the analysis plan in advance and apply Benjamini-Hochberg across the pre-registered family \cite{benjamini1995fdr}.

\paragraph{L14. Statistical power.} With $n$ in the low thousands and effect sizes of a few balanced-accuracy points, power is limited (\S\ref{sec:phase0}), and underpowered designs produce unreliable literatures \cite{card2020power,dror2018hitchhiker}. Power should be computed from the calibration curve before collection; a study that cannot reach $0.8$ at the target $m$ should say so rather than run.

\paragraph{L15. Dual use.} A reliable membership detector applied to individuals' text is a privacy attack \cite{carlini2021extracting,nasr2023scalable}. The set-level framing limits this, but item margins are individual scores and the calibration would transfer. Released artefacts should be restricted to benchmark items and public-corpus documents.

\section{Phase 1: the Protocol on Real Models}
\label{sec:phase1}

Everything to this point is either a proposal or a measurement of a procedure
on synthetic data. This section reports the protocol run against real
transformers: \PoneN{} audits over \PoneModels{} Pythia scales and two sources,
about an hour of compute on a laptop. It is a small study and we draw
correspondingly narrow conclusions, but one of them settles a question
\S\ref{sec:phase0} could not.

\input{generated/phase1_table.tex}

\subsection{Surface separability of published splits}

The cheapest thing the protocol produces is $\BA_{\nuis}$, the accuracy of a
classifier that never queries the model. On WikiMIA it is \PoneWkNuisLo{} to
\PoneWkNuisHi{}. The two sets are a temporal split, so they differ in topic and
vocabulary, and a blind classifier separates them almost perfectly. This
reproduces the finding of \citet{das2024blind} through a different instrument
and makes its consequence concrete: no statistic computed on that split, ours
included, can be read as evidence about memorisation. The protocol refuses the
verdict rather than reporting it (\PoneBlocked{} of \PoneN{} audits).

Pile train against Pile validation gives \PonePileNuisLo{} to
\PonePileNuisHi{}, which is chance. For the two subdomains we looked at, at
\PoneNsuspect{} items per set, construction E3 delivers what it promises. That
is worth recording because \citet{das2024blind} document drift in some Pile
subdomains, and it is not universal.

\subsection{Real depth profiles are not flat, and the correction scales}

\begin{figure*}[t]
\centering
\includegraphics[width=0.96\textwidth]{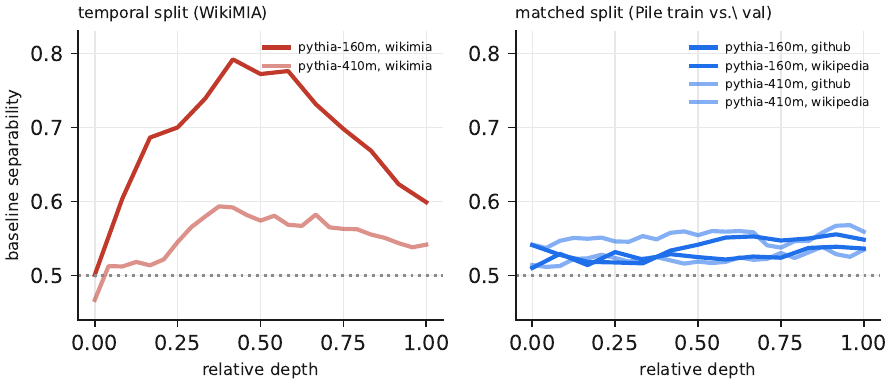}
\caption{Baseline depth profiles measured on real models. Each curve is the
separability of two halves of a reference set, where every item is a
non-member, so it is the profile of \emph{surface} separability in that model.
The flat-profile assumption that \S\ref{sec:phase0} had to discard is the
claim that these are horizontal lines.}
\label{fig:p1-profiles}
\end{figure*}

\S\ref{sec:phase0} showed on synthetic data that a contrast against a flat
depth profile fails in both directions, and could not say which way real
transformers go. Figure~\ref{fig:p1-profiles} answers it. The profiles are not
flat. On the temporal split they span \PoneWkSpanLo{} to \PoneWkSpanHi{}
accuracy points and are hump-shaped, rising steeply to a mid-network peak and
falling away. A test built on the assumption that surface decodability is
constant in depth would be comparing against a line that is wrong by up to
\PoneWkSpanHi{} points.

\begin{figure}[t]
\centering
\includegraphics[width=\columnwidth]{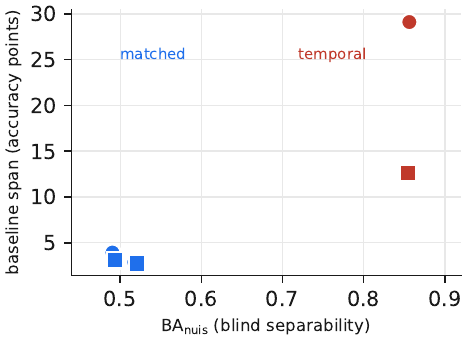}
\caption{The baseline correction scales with the surface difference it exists
to absorb. Circles are the 160M model, squares 410M.}
\label{fig:p1-span}
\end{figure}

The more useful finding is that the non-flatness tracks the surface
difference. Across the \PoneN{} audits the correlation between $\BA_{\nuis}$
and the baseline span is $\PoneCorr$ (Figure~\ref{fig:p1-span}). Where the two
sets are well matched the baseline is nearly flat, \PonePileSpanLo{} to
\PonePileSpanHi{} points, and the correction is close to a no-op. Where they
are not, the baseline is large and absorbs \PoneShareLo{} to \PoneShareHi{}
per cent of the raw statistic. A correction that is largest exactly where it is
most needed, and negligible where it is not, is the behaviour one wants; the
alternative would be a knob that moves results around indiscriminately.

\subsection{The negative control behaves}

All \PonePileNulls{} Pile arms return null, at both scales and on both
subdomains, with a negative adjusted statistic in every case. This is what the
literature predicts: Pythia sees the Pile approximately once, and
\citet{duan2024mia} found membership inference near chance in exactly this
setting. An instrument that manufactured positives on well-matched sets would
have shown it here, and did not.

The null has to be read at the right strength. Without the exposure
calibration of \S\ref{sec:programme} we cannot convert it into a statement
about duplication counts, so the correct reading is that no depth-dependent
familiarity signal is detectable at this sensitivity, not that none exists.

\subsection{A split where Requirement E holds by construction}
\label{sec:phase3}

Every arm above rests on an exchangeability argument. \citet{oren2024proving}
provide one that does not. They injected a random subset of PIQA's test file
into the pretraining corpus of a 1.4B model at a known duplication count and
released the whole file alongside the checkpoint, so the withheld items are
recoverable by difference. Both arms are then one pool split at random before
training, which is construction E1: exchangeability holds by randomisation
rather than by argument. All \PthreeNsuspect{} injected items used here are
present in the public file, which is the condition that makes the difference a
withheld arm rather than a second population, and the builder refuses if it
does not hold.

\input{generated/phase3_table.tex}

The measured $\BA_{\nuis}$ of \PthreeNuisLo{} to \PthreeNuisHi{} confirms the
construction empirically: a classifier with no access to the model cannot
separate the injected half from the withheld half. This is the only arm in the
paper where that is true, and it is worth contrasting with the
\PoneWkNuisLo{}--\PoneWkNuisHi{} of the temporal split.

We ran \PthreeN{} arms: two duplication counts, $m=\PthreeDupLo{}$ and
$m=\PthreeDupHi{}$, crossed with two prefix lengths, at a common sample size.
The two exposures use the same injected items and the same withheld items, so
only the exposure differs. The prefix arms differ only in how much of each
record the probe reads, because a null on a \PthreeWordsGoal{}-word prefix
would say more about the prefix than about the instrument, and the full record
is \PthreeWordsFull{} words.

\PthreeBlocked{} of the \PthreeN{} arms reaches nominal significance, at
$p=\PthreeBlockedP{}$, and it is the lower exposure. That ordering is already
wrong: more duplication should give more signal, not less. Two further facts
finish it off. Within that exposure the two prefix lengths give adjusted
statistics of opposite sign on identical items. And the significance comes
entirely from the recentring, since the uncorrected contrast on that arm is
null.

\paragraph{The correction carries more variance than the null.} Recentring
subtracts an estimate. The placebo baseline comes from a random split of the
reference set at a searched noise level, and the permutation null of
\S\ref{sec:method} describes the observed contrast with that estimate held
fixed. Re-estimating it over eight split seeds per arm gives a standard
deviation of \PthreeBaseSdLo{} to \PthreeBaseSdHi{}, against a null standard
deviation smaller than that in \PthreeRatioAboveOne{} of \PthreeN{} arms: the
ratio runs from \PthreeRatioLo{} to \PthreeRatioHi{}. The quantity subtracted
to remove a bias is more variable than the quantity it is subtracted from.

Propagating that variance turns $p=\PthreeBlockedP{}$ into
$p=\PthreeBlockedPinfl{}$, and the protocol issues no verdict. The apparent
detection was an artefact of which placebo split the search happened to draw:
across those eight seeds the baseline ranged widely enough that a different
draw would have moved the statistic by more than its own size.

This is a defect in the protocol as first specified, not in this dataset. It
does not touch the nulls, since an unmodelled variance component can only make
a null more null, and it does not undo \S\ref{sec:phase0}'s finding that
contrasting against a flat profile fails in both directions. The recentring is
still necessary. What it is not is free, and \S\ref{sec:basevar} states the
measurement that now has to accompany it.

Two things survive from the original reading. The null should be read at the
sensitivity \S\ref{sec:phase0} establishes, which is real but bounded, rather
than as an absence of any effect. And it extends the calibration point of
Table~\ref{tab:we2published}, where every published method including an
activation probe sits within a few points of chance at $m\in\{1,2\}$, out to
duplication counts of \PthreeDupLo{} and \PthreeDupHi{}.

\begin{figure*}[t]
\centering
\includegraphics[width=0.96\textwidth]{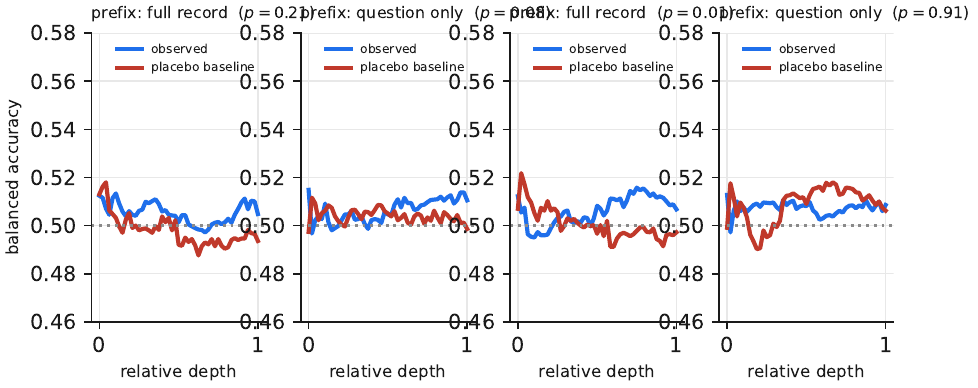}
\caption{Phase 3, measured. Observed depth profile against the level-matched
placebo baseline on \citeauthor{oren2024proving}'s contaminated 1.4B
checkpoint, PIQA injected at $m=\PthreeDup{}$, under two prefix lengths at the
same sample size. The baseline is not flat in either panel, and its range
scales with prefix length, which is why it has to be measured rather than
assumed. Neither observed profile departs from it significantly.}
\label{fig:phase3}
\end{figure*}

The baseline profile is again not flat, spanning \PthreeSpanLo{} to
\PthreeSpanHi{} accuracy points, and its range scales with prefix length: the
wider span belongs to the full record (Figure~\ref{fig:phase3}). Both arms
peak in the first few layers. Surface separability is therefore a property of
the inputs as well as of the model, which is a further reason its depth
profile cannot be assumed.

We report this comparison at a single sample size deliberately. An earlier
pairing ran the short prefix at half the items and appeared to show the
profile changing \emph{shape}, falling rather than humped. That difference did
not survive matching the sample sizes, and reporting it would have been the
same error this paper is about: reading a difference off two arms that
differed in more than one respect.

\subsection{What this does not show}

The two significant contrasts are on the split where Requirement E fails, and
they are therefore uninterpretable as contamination evidence. WikiMIA members
are pre-cutoff Wikipedia text and plausibly are in Pythia's training data, so a
true positive is entirely possible; the temporal confound is equally possible;
and nothing in the data distinguishes them. That is the point of the gate.

The Phase 3 null is the more informative result, and it should not be
overstated either. It is one benchmark, one model, one duplication count, and
a null at $m=\PthreeDup{}$ does not establish that no familiarity direction
exists at any exposure. Converting it into a statement about duplication
counts needs the exposure calibration of \S\ref{sec:programme}, which requires
controlling training. What it does establish is that a linear probe on the
residual stream, evaluated against a level-matched baseline on a split where
exchangeability is exact, finds nothing on a model that demonstrably memorised
the items. Read together with \S\ref{sec:phase1}, the pattern is that the
published positives live on splits that cannot support them, and the splits
that can support a verdict return nothing.

\section{Reproducibility and Artefacts}
\label{sec:repro}

Every number in \S\ref{sec:phase0} is written into this document by a
generated macro file rather than transcribed, so the paper cannot drift from
the run that produced it. The released artefacts are:

\paragraph{What the protocol costs.} Worth stating, because it bounds what can
be audited. Cross-fitting produces a dense $N\times N$ smoother per layer per
fold seed, so the memory is $O(\text{seeds}\cdot N^{2})$ if the layers are
streamed and $O(\text{layers}\cdot\text{seeds}\cdot N^{2})$ if they are
retained. At the \PthreeLayers{} layers and $N=3000$ of \S\ref{sec:phase3},
with five seeds, that is $0.36$\,GB against $17.6$\,GB. The released
implementation streams: the permutation null consumes each layer's smoothers
once, so accumulating the weighted contribution and releasing them costs
nothing and is what makes deep models tractable at all. Compute scales as
$O(\text{layers}\cdot\text{seeds}\cdot N^{2}d)$ for the fits and
$O(\text{layers}\cdot\text{seeds}\cdot N^{2}B)$ for the null; with activations
cached, the statistical stage for \PthreeLayers{} layers at $N=3000$ takes
about seven minutes on an M-series laptop. The dominant cost of an audit is
the forward passes, not the statistics.

\begin{itemize}\itemsep1pt
\item \texttt{rscp.py}, the protocol, which imports and runs on numpy alone;
      torch is needed only to extract activations from a real model.
\item \texttt{simulators.py}, both generative models, with the structural
      differences of Table~\ref{tab:sims} computable in one command.
\item \texttt{phase0c\_validation.py}, the study reported here, checkpointed
      per cell so it can be resumed.
\item \texttt{test\_rscp.py}, 29 property tests covering the claims the method
      rests on: that the smoother is label-independent and strictly
      out-of-fold, that its predictions are linear in the labels, that the
      contrast cancels any constant added to every layer, that the placebo
      refuses a degenerate surface key rather than returning a value, and that
      a planted effect of known size is recovered.
\item \texttt{run\_rscp\_eval.py}, the audit runner, with a dry-run mode that
      exercises the whole path without a GPU.
\item A \texttt{Makefile} whose \texttt{all} target runs the tests, reruns the
      study, regenerates every figure and table, and rebuilds this document.
      It takes about ten minutes on four CPU cores and needs no model weights.
\end{itemize}

Two of the bugs fixed during development were found by the dry run rather than
by inspection: a degenerate surface key produced a non-finite statistic, and
that non-finite value was being compared into a \emph{significant} placebo,
which would have silently blocked legitimate audits. Both are now covered by
tests. We mention it because a contamination detector that fails closed
without saying so is worse than no detector.

\section{Follow-on Empirical Validation Programme}
\label{sec:programme}

This section is the validation plan, not a report of results. Nothing in \S\ref{sec:worked} has been run.

\subsection{Pre-registered hypotheses}

\begin{itemize}\itemsep2pt
\item[\textbf{H1}] \emph{Nuisance dominance.} On temporally split membership benchmarks, $\BA_{\nuis}\ge 0.65$, and $\Delta_{\ell}$ is small relative to $\BA_{\nuis}-0.5$. Falsified if $\BA_{\nuis}<0.55$ on WikiMIA-style splits.
\item[\textbf{H2}] \emph{Exposure monotonicity.} Under randomised injection, $T$ increases monotonically in $m$ and is fit by Eq.~\eqref{eq:link} with $R^2\ge0.8$. Falsified by a non-monotone or flat curve.
\item[\textbf{H3}] \emph{Layer profile.} The depth contrast is positive and $\ell^{*}$ falls in the middle-to-late third, not at $\ell\le 2$. Falsified if $\ell^{*}$ concentrates in the first two layers, which would indicate the signal is embedding-level and hence nuisance that $\Phi_0$ failed to capture.
\item[\textbf{H0}] \emph{Depth profile of surface separability.} On real models the placebo baseline is not flat, and its shape differs across models and item pools. This is the assumption \S\ref{sec:phase0} had to remove rather than defend, and measuring the actual shapes is the first thing Phase 1 delivers. Falsified if placebo profiles turn out flat, in which case the recentring is unnecessary machinery.
\item[\textbf{H4}] \emph{Extraction-point separation.} Full-item extraction yields substantially larger apparent $T$ than prefix extraction on item sets constructed to differ in difficulty but not in membership. Falsified if the two extraction points agree on such sets, which would remove the motivation for the prefix rule.
\item[\textbf{H5}] \emph{Recency decay.} On Pythia, $T$ decreases with training steps elapsed since exposure. Falsified by a flat profile.
\end{itemize}

H3 and H4 are the two that carry the paper's specific methodological claims. H2 determines whether calibration is possible at all.

\subsection{Phases}

\textbf{Phase 0, procedure validation. Done; reported in \S\ref{sec:phase0}.} It falsified three of our own design choices, promoted the placebo from a veto to the baseline, and added the $|R| = 2|S|$ requirement.

\textbf{Phase 1, nuisance audit. Done; reported in \S\ref{sec:phase1}.} It measured real baseline depth profiles for the first time, confirmed the flat-profile assumption is violated, and returned null on every well-matched control.

\textbf{Phase 1, nuisance audit of existing benchmarks.} Compute $\BA_{\nuis}$ for WikiMIA, for the arXiv-based split of \citet{liu2024probing}, for Pile train versus validation per subdomain, and for GSM8K versus GSM1k. This phase requires forward passes only and produces a standalone result: a table of how much of each existing evaluation split is blind-separable. It also directly extends \citet{das2024blind} to splits they did not cover.

\textbf{Phase 2, calibration by injection.} Continued pretraining of Pythia-160M and Pythia-410M on a held-out Pile slice with randomised injection at $m\in\{0,1,2,4,8,16,32\}$ across the four format conditions of \S\ref{sec:we2}. Fit Eq.~\eqref{eq:link}; report $m^{*}$ and the saturation point. This is the compute-dominant phase.

\textbf{Phase 3, public contaminated checkpoints.} Done, and reported in \S\ref{sec:phase3}: RSCP on the checkpoints of \citet{oren2024proving} at $m=\PthreeDup{}$, the highest duplication count their release provides, directly comparable to the published AUCs in Table~\ref{tab:we2published}. The remaining work here is the lower duplication counts and the smaller scales, which their release also covers.

\textbf{Phase 4, application.} GSM8K versus GSM1k across the three models of \S\ref{sec:we3}, with the OLMo corpus search as partial ground truth.

\textbf{Phase 5, robustness.} Ridge versus logistic probes; $(\mathrm{pre},\mathrm{last})$ versus $(\mathrm{pre},\mathrm{mean})$; sensitivity to $n$; the pruned-benchmark question of \S\ref{sec:discussion}; and the scale trend across the Pythia suite.

\subsection{Compute budget}

The figures below are estimates derived from parameter counts and token counts, not measurements.

Activation extraction is negligible: a forward pass over 4000 items of roughly 200 tokens on a 1.4B model is on the order of $10^{15}$ floating-point operations, under a minute on a consumer GPU, with a few hundred megabytes of cache per model. Inference is seconds per configuration, as \S\ref{sec:phase0} measured.

Phase 2 dominates. Continued pretraining of Pythia-410M on $2\times10^{8}$ tokens costs roughly $6ND\approx 5\times10^{17}$ operations; at an effective $3\times10^{13}$ operations per second that is about five hours, and two hours at 160M. Seven exposure levels times four format conditions times two model sizes is a large grid, so we recommend the full grid at 160M and a reduced grid at 410M, for an estimated 60 to 80 GPU-hours, roughly 100 for the programme, a fortnight on one consumer card. Phases 1, 3 and 4 need forward passes only and run on Apple silicon; Phase 2 wants a discrete GPU.

\subsection{Artefacts}

Release the cached activations for all public-checkpoint configurations, the fitted probe weights and nuisance weights, the permutation null distributions rather than only the $p$-values, the injection corpora and continued-pretraining checkpoints from Phase 2, and the analysis code with fixed seeds. The permutation nulls matter: they let others recompute a $p$-value under a different test statistic without rerunning the pipeline. Evaluation should use a standard harness so that accuracy numbers in \S\ref{sec:adjusted} are comparable to published ones \cite{gao2024harness,wolf2020transformers}.

\subsection{Decision rule}

If H2 holds and $m^{*}\le 4$ for the verbatim condition, RSCP is a usable auditing instrument and the follow-on work is deployment across benchmarks with commissioned twins. If H2 holds but $m^{*}$ is large, the contribution is the detection floor itself, reported as a bound on what activation probing can achieve. If H2 fails, the honest conclusion is that representational familiarity is not a stable function of exposure count at these scales, and Phase 1's nuisance audit is then the paper's contribution while the probe is not.

\section{Conclusion}

Probing the residual stream is a reasonable way to look for benchmark
contamination, and the obvious way to do it does not work. Reporting the level
of a probe's advantage over a control makes the false positive rate a function
of how large the analyst made the control. Comparing the depth profile against
a flat expectation fails whichever way real profiles slope, and they slope both
ways. Resampling items with the probe held fixed understates the variance of
the statistic. Estimating the baseline on half as many items as the comparison
is not a baseline.

What is left after removing those is a protocol with four parts: read
activations at the item prefix so the detector cannot see the model's own
answer; contrast the depth profile rather than its level; recentre on a
placebo baseline built from the reference set at matched size and matched
embedding-layer separability; and test against a permutation null that refits
the probe. Its central move is to stop assuming what a null depth profile looks
like and measure it per model instead, using data the audit already requires.
Under two deliberately dissimilar simulators it holds its nominal error rate
where each of the discarded alternatives fails, and it reaches useful power at
separations of one to two accuracy points.

Phase 1 has now been run, and it supports the design without settling the
science. Real baseline profiles are not flat, they are hump-shaped, and their
departure from flatness scales with how different the two item sets are as
text. The well-matched negative controls come back null at both scales. The
protocol declines to rule on the one split where a positive appeared, because
that split is a temporal one on which a blind classifier already reaches
\PoneWkNuisHi{}.

The honest summary of the evidence is therefore narrow. A procedure behaves
correctly on synthetic activations where several plausible alternatives do not;
it runs on real models; its central assumption was worth removing. Whether a
transformer carries a linearly decodable familiarity signal, and at what
duplication count, is untouched by any of it, because the only positive result
sits behind a failed exchangeability check. Settling that needs the injection
experiments of \S\ref{sec:programme} and a benchmark with a reference set that
qualifies.

\end{document}

%% file: generated/results_macros.tex
\newcommand{\PropAGrowth}{1.00}
\newcommand{\PropACorr}{0.01}
\newcommand{\PropAKurt}{3.0}
\newcommand{\PropAEig}{0.025}
\newcommand{\PropBGrowth}{1.75}
\newcommand{\PropBCorr}{0.65}
\newcommand{\PropBKurt}{9.3}
\newcommand{\PropBEig}{0.188}

\newcommand{\CapContrastMatch}{0.020}
\newcommand{\CapMaxMatch}{0.030}
\newcommand{\CapNaiveMatch}{1.000}

\newcommand{\CapContrastHuge}{0.025}
\newcommand{\CapMaxHuge}{0.988}

\newcommand{\BootMin}{0.030}
\newcommand{\BootMax}{0.087}

\newcommand{\CapPdimHi}{1200}

\newcommand{\SlopeARaw}{(0,0.01667) (0.25,0.2833) (0.5,0.7167)}

\newcommand{\SlopeAAdjZero}{0.083}

\newcommand{\SlopeARawFifty}{0.717}
\newcommand{\SlopeAAdjFifty}{0.050}

\newcommand{\SlopeBAdjZero}{0.083}

\newcommand{\PwAFifteen}{0.860}

\newcommand{\PwAThirty}{1.000}

\newcommand{\PwBFifteen}{0.780}

\newcommand{\ProfNullFirstB}{0.754}
\newcommand{\ProfNullLastB}{0.657}

\newcommand{\SizeMatchedA}{0.067}
\newcommand{\SizeHalfA}{0.217}

\newcommand{\KeyFprFive}{0.050}
\newcommand{\KeyPwFive}{0.825}

\newcommand{\KeyPwFifty}{0.450}
\newcommand{\TimeSetup}{6.51}
\newcommand{\TimePerPerm}{1.19}
\newcommand{\TimeB}{2.4}
\newcommand{\TimeRefitH}{1.9}
\newcommand{\TimeSpeedup}{2919}
\newcommand{\TimeN}{1000}
\newcommand{\TimeD}{2048}
\newcommand{\RunMinutes}{10}
\newcommand{\NPerSet}{200}
\newcommand{\BPerm}{300}

%% file: generated/phase1_macros.tex
\newcommand{\PoneN}{6}
\newcommand{\PoneModels}{2}
\newcommand{\PoneCorr}{0.87}
\newcommand{\PoneWkNuisLo}{0.855}
\newcommand{\PoneWkNuisHi}{0.856}
\newcommand{\PonePileNuisLo}{0.490}
\newcommand{\PonePileNuisHi}{0.521}
\newcommand{\PoneWkSpanLo}{12.6}
\newcommand{\PoneWkSpanHi}{29.1}
\newcommand{\PonePileSpanLo}{2.7}
\newcommand{\PonePileSpanHi}{3.9}
\newcommand{\PonePileNulls}{4}

\newcommand{\PoneBlocked}{2}
\newcommand{\PoneShareLo}{25}
\newcommand{\PoneShareHi}{28}
\newcommand{\PoneNsuspect}{500}

%% file: generated/phase3_macros.tex
\newcommand{\PthreeN}{4}

\newcommand{\PthreeDup}{50}
\newcommand{\PthreeDupLo}{7}
\newcommand{\PthreeDupHi}{50}
\newcommand{\PthreeNsuspect}{1000}
\newcommand{\PthreeLayers}{49}
\newcommand{\PthreeNuisLo}{0.508}
\newcommand{\PthreeNuisHi}{0.515}

\newcommand{\PthreeSpanLo}{1.4}
\newcommand{\PthreeSpanHi}{3.1}

\newcommand{\PthreeGapLo}{0.001}

\newcommand{\PthreeRatioLo}{1.30}
\newcommand{\PthreeRatioHi}{1.56}
\newcommand{\PthreeRatioAboveOne}{4}
\newcommand{\PthreeBaseSdLo}{0.00456}
\newcommand{\PthreeBaseSdHi}{0.00566}
\newcommand{\PthreeBlocked}{1}
\newcommand{\PthreeBlockedP}{0.0075}
\newcommand{\PthreeBlockedPinfl}{0.0745}
\newcommand{\PthreeWordsFull}{44}
\newcommand{\PthreeWordsGoal}{7}

%% file: generated/results_tables.tex

\begin{table}[t]
\small
\centering
\setlength{\tabcolsep}{4pt}
\begin{tabular}{@{}lccc@{}}
\toprule
Statistic & \multicolumn{3}{c}{control set dimension $p$} \\
\cmidrule(l){2-4}
 & 96 & 400 & 1200 \\
\midrule
raw accuracy, no control & 1.000 & 1.000 & 1.000 \\
$\max_\ell \Delta_\ell$ (level) & 0.030 & 0.530 & 0.988 \\
$\overline{\Delta_\ell}$ (level) & 0.030 & 0.640 & 0.975 \\
\addlinespace
\textbf{depth contrast} & \textbf{0.020} & \textbf{0.000} & \textbf{0.025} \\
\midrule
$\BA_{\nuis}$ & 0.652 & 0.609 & 0.570 \\
\bottomrule
\end{tabular}
\caption{\textbf{C1.} False positive rate under a true null as the nuisance control set grows, Sim-A, nominal $0.05$. The level of excess separability inherits the control set's estimation capacity; the zero-sum contrast cancels it algebraically.}
\label{tab:c1}
\end{table}

\begin{table}[t]
\small
\centering
\setlength{\tabcolsep}{5pt}
\begin{tabular}{@{}llcc@{}}
\toprule
Simulator & depth slope & uncorrected & recentred \\
\midrule
Sim-A & 0\% & 0.017 & \textbf{0.083} \\
 & 25\% & 0.283 & \textbf{0.050} \\
 & 50\% & 0.717 & \textbf{0.050} \\
\addlinespace
Sim-B & 0\% & 0.000 & \textbf{0.083} \\
 & 25\% & 0.000 & \textbf{0.083} \\
 & 50\% & 0.000 & \textbf{0.033} \\
\bottomrule
\end{tabular}
\caption{\textbf{C2.} False positive rate under a true null when surface decodability grows with depth. Recentring on the level-matched placebo holds the nominal rate where the uncorrected contrast does not.}
\label{tab:c2}
\end{table}

%% file: generated/phase1_table.tex
\begin{table*}[t]
\small
\centering
\setlength{\tabcolsep}{5pt}
\begin{tabular}{@{}llrrrrrrl@{}}
\toprule
Model & Split & $n_S$ & $\BA_{\nuis}$ & baseline span & $T_{\mathrm{raw}}$ & baseline & $T_{\mathrm{adj}}$ & Outcome \\
\midrule
pythia-160m & pile github & 500 & 0.490 & 3.9 pts & -0.0099 & +0.0116 & -0.0215 & null, $p=1.00$ \\
pythia-410m & pile github & 500 & 0.494 & 3.1 pts & -0.0067 & +0.0048 & -0.0114 & null, $p=0.96$ \\
pythia-160m & pile wikipedia & 500 & 0.517 & 2.9 pts & -0.0040 & +0.0066 & -0.0106 & null, $p=0.94$ \\
pythia-410m & pile wikipedia & 500 & 0.521 & 2.7 pts & -0.0048 & +0.0050 & -0.0098 & null, $p=0.94$ \\
pythia-160m & wikimia & 394 & 0.856 & 29.1 pts & +0.0498 & +0.0138 & +0.0361 & no verdict, Req.\ E fails \\
pythia-410m & wikimia & 394 & 0.855 & 12.6 pts & +0.0610 & +0.0153 & +0.0457 & no verdict, Req.\ E fails \\
\bottomrule
\end{tabular}
\caption{Phase 1, measured. $\BA_{\nuis}$ is what a classifier with no access to the model achieves; near $0.5$ the two sets are exchangeable, near $0.86$ they are not and no contamination verdict is admissible. The baseline span is the range of the placebo depth profile, and it is the quantity the flat-profile assumption gets wrong. Every Pile arm is null with a negative adjusted statistic, which is what the literature predicts for a corpus seen approximately once.}
\label{tab:phase1}
\end{table*}

%% file: generated/phase3_table.tex
\begin{table}[t]
\small
\centering
\setlength{\tabcolsep}{4pt}
\begin{tabular}{@{}rlrrrrrl@{}}
\toprule
$m$ & Prefix & $\BA_{\nuis}$ & $T_{\mathrm{adj}}$ & $p$ & $\frac{\mathrm{sd(base)}}{\mathrm{sd(null)}}$ & $p^{*}$ & Outcome \\
\midrule
50 & full record & 0.515 & +0.0035 & 0.210 & 1.30 & 0.312 & null \\
50 & question only & 0.508 & +0.0049 & 0.080 & 1.56 & 0.221 & null \\
7 & full record & 0.514 & +0.0108 & 0.007 & 1.31 & 0.074 & no verdict \\
7 & question only & 0.509 & -0.0046 & 0.909 & 1.30 & 0.803 & null \\
\bottomrule
\end{tabular}
\caption{Phase 3, measured. \citeauthor{oren2024proving}'s contaminated 1.4B checkpoints, PIQA injected at two duplication counts. Both arms are the same test file split at random before training, so Requirement E holds by construction and $\BA_{\nuis}$ confirms it. $p$ holds the placebo baseline fixed, as the protocol originally specified; $p^{*}$ propagates the baseline's own sampling variance, measured over eight split seeds. The ratio column is that variance against the null's. It exceeds one in every arm, which is why the one nominally significant result carries no verdict.}
\label{tab:phase3}
\end{table}